%% file: main.tex
\documentclass[authoryear,preprint,12pt]{elsarticle}

\usepackage{amssymb}
\usepackage{amsthm}
 \usepackage{listings}
\usepackage{color}
\usepackage{stfloats}
\usepackage[caption=false,font=footnotesize]{subfig}
\usepackage{svg}
\usepackage{amsmath}
\RequirePackage{multirow}

\allowdisplaybreaks[1]

\usepackage{blindtext}
\usepackage{hyperref}
\hypersetup{
    colorlinks=true,
    linkcolor=blue,
    filecolor=magenta,      
    urlcolor=cyan,
    pdftitle={Overleaf Example},
    pdfpagemode=FullScreen,
    }

\usepackage{xcolor}

\definecolor{codegreen}{rgb}{0,0.6,0}
\definecolor{codegray}{rgb}{0.5,0.5,0.5}
\definecolor{codepurple}{rgb}{0.58,0,0.82}
\definecolor{backcolour}{rgb}{0.95,0.95,0.92}

\lstdefinestyle{mystyle}{
    backgroundcolor=\color{backcolour},   
    commentstyle=\color{codegreen},
    keywordstyle=\color{magenta},
    numberstyle=\tiny\color{codegray},
    stringstyle=\color{codepurple},
    basicstyle=\ttfamily\footnotesize,
    breakatwhitespace=false,         
    breaklines=true,                 
    captionpos=b,                    
    keepspaces=true,                 
    numbers=left,                    
    numbersep=5pt,                  
    showspaces=false,                
    showstringspaces=false,
    showtabs=false,                  
    tabsize=2
}

\usepackage{tikz}
\usetikzlibrary{arrows,arrows.meta,calc,positioning}
\usetikzlibrary{shapes,shapes.multipart,shapes.geometric}

\newcommand{\bb}[1]{\textcolor{black}{#1}}

\begin{document}

\begin{frontmatter}



\title{Physics-Informed Neural Networks with Skip Connections for Modeling and Control of Gas-Lifted Oil Wells}

 \author[NTNU]{Jonas Ekeland Kittelsen}
\author[UFSC]{Eric Aislan Antonelo}
\author[UFSC]{Eduardo Camponogara}
\author[NTNU]{Lars Struen Imsland}


 \affiliation[NTNU]{organization={Norwegian University of Science and Technology},
             city={Trondheim},
             country={Norway}}
\affiliation[UFSC]{organization={Federal University of Santa Catarina},
            city={Florianopolis},
            country={Brazil}}

\begin{abstract}
\bb{Neural networks, while powerful, often lack interpretability. Physics-Informed Neural Networks (PINNs) address this limitation by incorporating physics laws into the loss function, making them applicable to solving Ordinary Differential Equations (ODEs) and Partial Differential Equations (PDEs). 
The recently introduced PINC framework extends PINNs to control applications, allowing for open-ended long-range prediction and control of dynamic systems.
In this work, we enhance PINC for modeling highly nonlinear systems such as gas-lifted oil wells. By introducing skip connections in the PINC network and refining certain terms in the ODE, we achieve more accurate gradients during training, resulting in an effective modeling process for the oil well system.
 Our proposed improved PINC demonstrates superior performance, reducing the validation prediction error by an average of 67\% in the oil well application and significantly enhancing gradient flow through the network layers, increasing its magnitude by four orders of magnitude compared to the original PINC.
Furthermore, experiments showcase the efficacy of Model Predictive Control (MPC) in regulating the bottom-hole pressure of the oil well using the improved PINC model, even in the presence of noisy measurements.
}
\end{abstract}




\end{frontmatter}



\section{Introduction} \label{sec:intro}
Physics-Informed Neural Networks (PINNs) \citep{Raissi2019} have been widely used to solve Partial Differential Equations (PDEs) and Ordinary Differential Equations (ODEs), representing an alternative to numerical methods that can speed up simulation orders of magnitude. 
Once trained, these networks can output directly the solution without access to the PDE or ODE equations. Actually, these equations are employed only during the training phase by designing a loss function that considers a measure of the deviation of the derivative of the network's output from the physics laws given by ODEs/PDEs. In this way, the PINN is trained so that its output satisfies the constraints of the physics, in addition to the traditional data loss for regression, \textit{i.e.}, the mean squared error of the data points. Regarding an ODE or PDE, this data loss corresponds to the error concerning the initial or boundary conditions. For industrial plants represented by ODEs, one could collect more data points in addition to initial conditions to complement differential equations with uncertain or unknown parameters \citep{Raissi2019}.

Applications of PINNs are widespread in many engineering areas.
\cite{Ouyang2023} employed multiple serial PINNs to decouple the governing physics equations for learning the reconstruction of hydrofoil cavitation flow.
\cite{Nazari2023} successfully applied PINNs for modeling river channels and parameter discovery under conditions of limited system measurements
and \cite{Navier2022} employed PINNs for fluid flow generation.
Many extensions of PINNs exist. For instance, \cite{Dwivedi2020} extended Extreme Learning Machines as physics-informed networks for rapidly learning solutions to PDEs; 
\cite{Xiang2022} introduced an improved PINN with a self-adaptive loss function to dynamically weigh the loss terms relating to data and physics; 
\cite{Tang2023} devised a parallel physics-informed neural network to solve the multi-body dynamic equations in full-scale train collision simulation, while \cite{Zhou2022} developed a physics-informed generative adversarial network-based approach to facilitate uncertainty quantification and propagation in addition to enabling measurement data fusion into system reliability assessment.

PINNs have been extended for long-range simulation and control applications in \cite{PINC_paper}, as they cannot inherently cope with changing inputs or extrapolate the predictions for time periods longer than the one defined during training.   
To address these limitations, \cite{PINC_paper} presented a new framework called Physics-Informed Neural Nets for Control (PINC), which is a novel PINN-based architecture suitable for control problems and able to simulate for longer-range time horizons that are not fixed beforehand during training. 
As a result, PINC is more flexible than traditional PINNs and faster than numerical methods as it relies only on signal propagation through the network, making it less computationally costly and better suited for model simulation in Model Predictive Control, for instance. 
Model Predictive Control (MPC) is a widely applied method for multivariate control in both industry and academia \citep{Camacho2013}, which has been used successfully in various fields such as oil and gas \citep{Jordanou2021}, aerospace \citep{aerospace_mpc}, process industries, and robotics \citep{mpc_robotics} since its inception in the 1970s.

The class of systems that \cite{PINC_paper} showed PINC to be effective include the Van der Pol oscillator and the four-tank system. 
  \bb{Despite being benchmark systems for nonlinear dynamics, their dynamic equations are smooth functions that may lack specific characteristics of real-world complex systems. These particular features such as nonsmoothness may hinder the training of PINC networks.}
Here, we investigate improved versions of the PINC framework for a class of systems that exhibit highly complex and nonsmooth behavior. In particular, we are interested in modeling and control of gas-lifted oil wells, which are discussed in Section~\ref{sec:oilwell}. Applying PINC as proposed in its original version for modeling and controlling oil wells considered in this paper did not succeed. There are three main reasons for that: 1) the magnitude of the gradient of the physics-loss function was too low for training to progress consistently; 2) some ODE equations of the well model include functions such as square root and logarithm, which are not defined for negative numbers; 3) and highly nonlinear terms of the ODE can negatively influence the training of the PINC. 
\bb{Thus, the complexity of the class of systems we consider arguably comes from these nonlinear terms, for instance, involving square roots, fractional exponents, ratios, and maximum operators. They can introduce nonsmooth points in mathematical functions, creating numerical problems when computing derivatives.
Some of these systems may exhibit stiffness due to the interaction of multiple physical processes that operate at different time scales. Also, nonsmoothness can originate from using max operators in ODEs.
}

The contributions of this work are discussed next. The first issue presented above was mainly tackled by using skip connections as in \cite{improved_NN_struct}, aiming to improve the magnitude of the gradient through the PINN's layers. In that work, they showed that traditional PINNs have specific gradient pathologies, which can be dealt with by the novel architecture and a learning rate annealing algorithm. In the current work, we transfer their proposed novel PINN architecture with skip connections to the PINC framework, showing that this improvement was essential for the modeling and controlling more complex systems such as the oil well.
The remaining issues, which refer to the terms in the ODE equations, are dealt with by modifying these equations when they appear in the physics-loss function for training the PINC. This is to say that surrogate terms can replace specific terms to render training smoother and more efficient. Notice that the latter procedure is specific to the application at hand, \textit{i.e.}, the oil well model. Lastly, we propose a hierarchical architecture with two modules: the first one corresponds to a PINC trained to predict the states of the oil well; and the second one is a standard feedforward neural network (NN) trained to predict the algebraic variables of the system (\textit{e.g.}, bottom-hole pressure) based on the predictions of the states by the PINC (first module), \textit{i.e.}, the output of the PINC is the input to the second NN. This NN streamlines the training of additional algebraic variables without retraining the PINC, which would be far more computationally costly than just training an NN.

\bb{The remainder of this work is organized as follows.
Section \ref{sec:related-works} discusses some works from the literature related to applying the PINC framework in model predictive control.
  Section \ref{sec:oilwell} introduces gas-lifting in oil wells and presents the differential algebraic equations that model the process. 
The PINN and PINC frameworks, along with the proposed improvements to the PINC, evaluation metrics, and MPC, are presented in Section~\ref{sec:methods}. In the following section, the experimental setup is described, while in Section~\ref{sec:results}, the results are showcased. Section~\ref{sec:conclusion} concludes this work.}

\section{Related Works}\label{sec:related-works}

\cite{Nicodemus2022} used the PINC framework, without skip connections (unlike our work), for model predictive control of multi-body dynamics. The results show that the PINC framework effectively solves a tracking problem for a complex mechanical system, a multi-link manipulator.
Applying to a PowerCube serial robot, they demonstrate
that the simulation of the nonlinear dynamics with PINC speeds up the computation time relative to numerical methods while retaining sufficient accuracy.

\cite{Liu2021} proposed a model-based Reinforcement Learning (RL) algorithm that utilizes physical laws to learn the state transition dynamics of an agent's environment. The algorithm employs an encoder-decoder recurrent network architecture to model the state transition function by minimizing the violation of conservation laws. The learned transition model is then employed to generate samples for an alternative replay buffer, which improves sample efficiency in the RL update process and reduces the need for real-world interactions. However, this approach differs from our PINC-based approach. \cite{Liu2021} build the physics-loss function on the discretized form of the laws of the system, unlike our work that relies on the continuous form. While they use recurrent networks, our approach employs feedforward networks in which time is explicitly given as input.

\cite{Gokhale2022} applied a PINN approach to learning a control-oriented thermal model of a building. The authors assume that the model is a discrete-time transition function in a Markov Decision Process (MDP) that predicts the next state, given the current state and action. This approach allows for control actions to be input to the model. However, unlike in our approach, their physics loss function also needs to be discretized. Although their proposal is control-oriented, they do not demonstrate actual control experiments with the trained PINNs, unlike the results from our work presented in Section~\ref{sec:results}.

\section{Gas-lifted Oil Wells} 
\label{sec:oilwell}

Primary recovery is the first stage of extracting oil and gas from reservoirs.
   It relies on the natural difference in pressure between the surface and the underground reservoir, therefore requiring relatively limited capital investment. 
With the gradual extraction of oil from the well, the pressure underground will slowly decrease, causing the volume of oil production to decline. Secondary recovery techniques are employed to mitigate the pressure loss, such as water injection, which seeks to force oil to the surface by directly applying pressure.
  Also, artificial lifting techniques are applied in wells, particularly those found in deep-water offshore reservoirs, to increase the pressure gradient from the bottom hole to the surface.
A widely used technique for artificial lifting is gas-lift. It works by injecting high-pressure gas at the bottom of the well, reducing the density of the production stream and forcing the flow of fluids to the surface. Gas-lift is a favored technique for its desirable features, which include low installation and maintenance cost, robustness, and a wide range of operating conditions.

 \begin{figure}[htb!]
     \centering
     \includegraphics[width=0.65\textwidth]{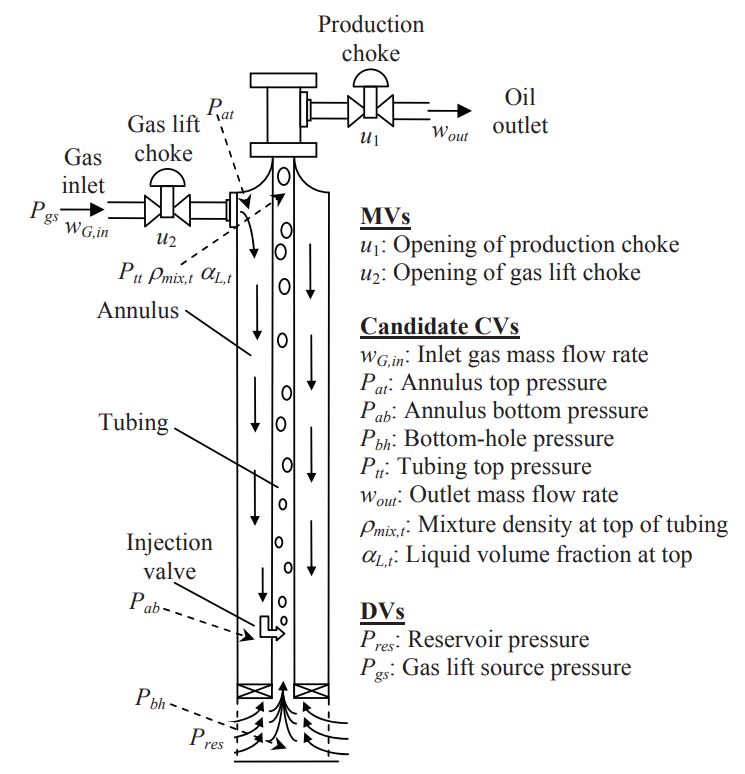}
     \caption{Schematics of a gas-lifted oil well \citep{well_model_paper:2012}.
      Liquid and gas from the reservoir enter the well at the bottom of the tubing, referred to as the bottom hole. From there, the gas and liquid flow up through the tubing until they exit through the production choke. The tubing is encased by a larger tube called the annulus. At the top of the annulus, gas is injected through the gas-lift choke. This gas flows down the annulus and enters the production tubing through a check valve (directional/one-way valve) near the annulus's bottom. This gas helps ``lift'' the reservoir flow up the tubing, increasing the production of the well. This kind of artificial lift is commonly used when the pressure in the reservoir is insufficient to sustain the flow from the reservoir to the top side.
     }
     \label{fig:well-structure}
 \end{figure}

The dynamic model for gas-lifted oil wells considered here is a system of  Differential Algebraic Equations (DAE) developed by \cite{well_model_paper:2012}. 
    The derivation of the model simplified some algebraic equations to avoid implicit terms, resulting in an explicit model. Consequently, the DAE system can be regarded as a system of Ordinary Differential Equations (ODE) and simulated using, for instance, the Runge-Kutta method.  
   These simplifications are imposed on the equations for velocities and will be explained further in the text.

\subsection{Well Model}

The model has several variables, several of which are similar in nature (pressures and densities, e.g.) but for different well locations. Subscript abbreviations are used extensively in the variables to refer to the phase (gas or liquid) and location. The variables are listed in Table~\ref{tab:model_abbreviations}. An important remark is that this model does not distinguish between water and oil, considering only a liquid phase.

\begin{table}[htb!]
    \centering
        \caption{Abbreviations used in the subscript of the model variables indicate the phase and location the respective variable represents.}
    \label{tab:model_abbreviations}
    \begin{tabular}{cc}
    \hline
         Abbreviation & Description \\ \hline
         $\tt G$ & gas \\
         $\tt L$ & liquid \\
         $\tt an$ & annulus \\
         $\tt tb$ & tubing \\
         $\tt bh$ & bottom-hole \\
         $\tt t$ & top \\
         $\tt b$ & bottom \\
         \hline
    \end{tabular}

\end{table}

\subsubsection{States}
The system state is characterized by the mass of gas in the annulus $m_{{\tt G},{\tt an}}$, mass of gas in the tubing $m_{{\tt G},{\tt tb}}$, and mass of liquid in the tubing $m_{{\tt L},{\tt tb}}$. The differential equations are:
\begin{subequations}
\begin{align}
        &\dot{m}_{{\tt G},{\tt an}} = w_{{\tt G},{\tt in}} - w_{{\tt G},{\tt inj}} \label{eq:well_ode_1}\\
        &\dot{m}_{{\tt G},{\tt tb}} = w_{{\tt G},{\tt inj}} + w_{{\tt G},{\tt res}} - w_{{\tt G},{\tt out}} \label{eq:well_ode_2}\\
        &\dot{m}_{{\tt L},{\tt tb}} = w_{{\tt L},{\tt res}} - w_{{\tt L},{\tt out}} \label{eq:well_ode_3}
\end{align}
\label{eq:well_ode}
\end{subequations}
where: $w_{{\tt G},{\tt in}}$ is the gas mass flow injected into the top of the annulus through the gas-lift choke; $w_{{\tt G},{\tt inj}}$ is the gas mass flow from the annulus into the tubing through the injection check valve, located close to the bottom of the well; $w_{{\tt G},{\tt res}}$ and $w_{{\tt L},{\tt res}}$ are the gas and liquid mass flows from the reservoir into the tubing; $w_{{\tt G},{\tt out}}$ and $w_{{\tt L},{\tt out}}$ are the gas and liquid mass flows out of the well through the production choke, or simply the gas and liquid production of the well.

The equations that calculate the flows appearing in state equations as well as the parameters for three oil well models used in the experiments are introduced in \ref{app:well}. 
Refer to \cite{well_model_paper:2012} for a full account of the model. The parameters appearing in the equations are explained in Table~\ref{tab:model_constants}.


\section{Methods}
\label{sec:methods}

\subsection{Physics-Informed Neural Networks (PINN)}

Physics-Informed Neural Networks (PINNs) \citep{Raissi2019} are trained to reproduce the solution of a dynamic system by utilizing the known (or partially known) underlying physics of the system. The use of physics knowledge greatly reduces the need for training data, in some cases removing it completely. 
Here, the output of the PINN represents the solution of an ODE. Let us consider the Initial Value Problem (IVP):
\begin{equation}
    \dot{\mathbf{y}}=\mathbf{f}(\mathbf{y}), \quad \mathbf{y}(0)=\mathbf{y}_0
    \label{eq:background_ODE}
\end{equation}
where we are interested in obtaining the solution $\mathbf{y}(t)$ on some interval $t \in \left [ 0, T \right ]$. 
   For some $\mathbf{f}(\mathbf{y})$, the IVP can be solved analytically, while for others, obtaining an analytic solution is not possible or practical, in which case we have the choice of using numerical methods, \textit{e.g.}, Runge-Kutta method. 
PINNs offer an alternative way of solving such problems. 

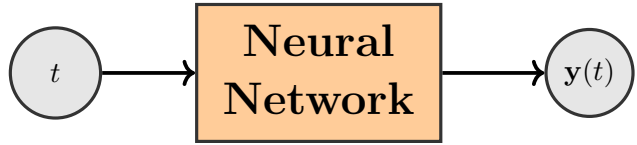
\begin{figure}[htb]
  \begin{center}
\begin{tikzpicture}[
orangenode/.style={rectangle, draw=black!80, fill=orange!40, very thick, minimum size=7mm},
greennode/.style={circle, draw=black!80, fill=green!30, very thick, minimum size=5mm},
graynode/.style={circle, draw=black!80, fill=black!10, very thick, minimum size=5mm},
]
\node[orangenode]      (explicit) at (0,0)   
        {\Large  \begin{tabular}{c} \textbf{Neural} \\ 
         \textbf{Network} \end{tabular}};
\node[graynode,minimum size=1.2cm] (t) at (-3.5,0) {\small $t$}; 
\draw[->]   [line width=0.5mm, black] (t) -- (explicit);

\node[graynode,minimum size=1cm] (y1) at (3.6,0) {\small $\mathbf{y}(t)$}; 
\draw[->]   [line width=0.5mm, black] (explicit) -- (y1);

\end{tikzpicture}
\end{center}
    \caption{PINN structure for solving an IVP. The neural network maps the time $t$ to the state at this time: $\mathbf{y}(t)$.}
    \label{fig:background_PINN_illustration}
\end{figure}

Consider the setup illustrated in Figure~\ref{fig:background_PINN_illustration}, where a neural network, with time $t$ as input, is trained to predict the corresponding states $\mathbf{y}(t)$ as output. 
    The network's training and prediction considers that $t$ is within the time horizon  $t \in \left [ 0, T \right ]$. 
In this conventional architecture, initial conditions are ``trained'' into the neural network weights, which means that the trained network will work only with the specified fixed initial conditions.
   Thus, we would have to train a new neural network if we want to solve an IVP for the same system but with a different initial condition $\mathbf{y}(0)$.

\subsubsection{Loss Function}
\label{sec:pinc_loss}
The loss function for PINNs consists of two terms basically:
\begin{equation}
    MSE = MSE_y + MSE_F
    \label{eq:background_PINN_MSE}
\end{equation}

While the first term, $MSE_y$, serves to impose the initial condition on the neural network by utilizing input data with corresponding output targets, the second term, $MSE_F$, is used to impose the physics of the system on the neural network output by penalizing it for not satisfying the ODE. 
   The latter term acts as a regularization term by imposing a specific solution on the neural network. In contrast, the goal of a traditional regularization term is usually to penalize large magnitudes of the weights.

\subsubsection*{Initial Conditions}
To impose the initial condition on the neural network, we create a training data point $(t,\hat{\mathbf{y}})$ consisting of an input  $t=0$ to the neural network and a corresponding target $\hat{\mathbf{y}}_0=\mathbf{y}_0$ for its output. 
Then, $MSE_y$ is calculated as follows:
\begin{equation}
    MSE_y = \frac{1}{N_y}  \left \| \mathbf{y}(0) - \hat{\mathbf{y}}_0 \right \|^2 
    \label{eq:background_PINN_MSE_y}
\end{equation}
  where:
$\mathbf{y}(0)$ is the predicted output of the neural network for input $t=0$;
   $N_y$ is the dimension of $\mathbf{y}$ or the number of system states;
and $\left \| \cdot \right \|$ is the $\ell_2$-norm.

\subsubsection*{Physics}

The neural network output should satisfy the ODE for the entire time horizon of interest $t \in \left [ 0, T \right ]$ for the network to adhere to the physics of the underlying system. For this, $N_f$ input points, $\{t^k: k=1,\dots,N_f\}$, are randomly sampled covering the entire time horizon $\left [ 0, T \right ]$, yielding a dataset of collocation points.  For each collocation point, a forward pass of the PINN is performed to calculate the corresponding output $\mathbf{y}(t^k)$. The resulting output is used to calculate the deviation from the ODE in a residual function named $\mathbf{F}$:
\begin{equation}
    \mathbf{F}(\mathbf{y}):= \frac{\partial \mathbf{y}}{\partial t} - \mathbf{f}(\mathbf{y})
    \label{eq:background_PINN_F}
\end{equation}
where the first term is the neural network output $\mathbf{y}(t)$ differentiated with respect to its time input. 
This derivative is obtained using automatic differentiation, for instance, using TensorFlow. 
The second term of this equation, $\mathbf{f}(\mathbf{y})$, corresponds to the ODE calculated for $\mathbf{y}$. 
If these two terms match, our neural network's solution $\mathbf{y}$ satisfies the system model given by the ODE.
 This residual function $\mathbf{F}$ is applied to every collocation point and used to calculate the physics-related loss $MSE_F$:
\begin{equation}
    MSE_F =  \frac{1}{N_f}\sum_{k=1}^{N_f} \frac{1}{N_y} \left \| \mathbf{F}(\mathbf{y}(t^k)) \right \| ^2
    \label{eq:background_PINN_MSE_f}
\end{equation}

The loss function defined in \eqref{eq:background_PINN_MSE} can now be computed and used for training the PINN. If it can reproduce the initial condition, and the derivative of the neural network output satisfies the ODE for the entire time interval, $\left [ 0, T \right ]$, then the output of the PINN is a solution to the IVP.

\subsubsection{Training}
\bb{
To train PINNs, we must minimize the total loss function in \eqref{eq:background_PINN_MSE}.
This minimization commonly employs a two-step strategy involving the Adam optimizer followed by the Broyden-Fletcher-Goldfarb-Shanno (BFGS) optimizer \citep{Fletcher2000}. 
The BFGS optimizer utilizes the Hessian matrix to determine the optimization direction, yielding more precise results by considering curvature in a high-dimensional space. However, if applied directly without preceding it with the Adam optimizer, there is a risk of rapid convergence to a local minimum for the residual without exploring other potential solutions. To mitigate this, we use the Adam optimizer initially to navigate away from local minima and subsequently refine the solution using the more accurate BFGS. In the context of Physics-Informed Neural Networks (PINNs), which often involve high-dimensional problems, the limited-memory version of BFGS (L-BFGS) is frequently employed to handle optimization problems with a large number of variables, as encountered in deep learning applications.
This combined approach, involving a sequence of optimizers, has been successfully applied in physics-informed neural networks, particularly for solving systems of partial differential equations (PDEs), as demonstrated in prior studies \citep{Karniadakis2021}. 
We apply the same methodology in the current study.
Additionally, \cite{Taylor2022} present a recent investigation on different optimizers for PINNs.
}

\subsection{Physics-Informed Neural Nets for Control (PINC)}

\begin{figure*}[htb]
\centering
\subfloat[PINC]{
          \label{fig:background_PINC_illustration}
\centering
\begin{tikzpicture}[
orangenode/.style={rectangle, draw=black!80, fill=orange!40, very thick, minimum size=7mm},
greennode/.style={circle, draw=black!80, fill=green!30, very thick, minimum size=5mm},
graynode/.style={circle, draw=black!80, fill=black!10, very thick, minimum size=5mm},
]
\node[orangenode,minimum height=4cm]      (nn) at (0,0)   
         {\Huge \textbf{NN} };
\node[graynode,minimum size=1.2cm] (t) at (-2.3,1.4) { $t$}; 
\draw[->]   [line width=0.5mm, black] (t.east) -- (nn);
  \node[graynode,minimum size=1.2cm] (y0) at (-2.3,0) {\small $\mathbf{y}(0)$}; 
   \draw[->]   [line width=0.5mm, black] (y0) -- (nn);
\node[graynode,minimum size=1.2cm] (u) at (-2.3,-1.4) {\small $\mathbf{u}$}; 
\draw[->,line width=0.5mm, black] (u.east) -- (nn);

\node[graynode,minimum size=1cm] (y1) at (2.3,0) {\small $\mathbf{y}(t)$}; 
\draw[->]   [line width=0.5mm, black] (nn) -- (y1);

\end{tikzpicture}
}
\subfloat[PINC in self-loop]{
          \label{fig:background_PINC_illustration_self_loop}
\centering
\begin{tikzpicture}[
orangenode/.style={rectangle, draw=black!80, fill=orange!40, very thick, minimum size=7mm},
greennode/.style={circle, draw=black!80, fill=green!30, very thick, minimum size=5mm},
graynode/.style={circle, draw=black!80, fill=black!10, very thick, minimum size=5mm},
]
\node[orangenode,minimum height=4cm]      (nn) at (0,0)   
         {\Huge \textbf{NN} };
\node[graynode,minimum size=1.2cm] (t) at (-2.3,1.4) { $t$}; 
\draw[->]   [line width=0.5mm, black] (t.east) -- (nn);
 
\node[graynode,minimum size=1.2cm] (u) at (-2.3,-1.4) {\small $\mathbf{u}$}; 
\draw[->,line width=0.5mm, black] (u.east) -- (nn);

\draw[->,line width=0.5mm, black] (nn.east) -- (2.2,0) -- (2.2,-2.8) -- (-3.5,-2.8) -- (-3.5,0) -- (nn);

\node (yT) at (1.55,0.3) {$\mathbf{y}(T)$};
\node (y0) at (-2.3,0.3) {$\mathbf{y}(0)$};

\end{tikzpicture}
}

    \caption{Physics-informed Neural Networks for Control (PINC). 
    Left: PINC network with time $t$, initial condition $\mathbf{y}(0)$, and control input $\mathbf{u}$ as inputs. This neural network can make predictions of the state at time $t$ given any initial condition and control input. 
    Right:
    PINC network in self-loop mode, allowing for long-range simulations. The output state prediction at the time $t=T$ is fed back as the initial condition of the PINC to make a new prediction, progressing $T$ seconds every iteration. A different control input may be applied at every iteration. For the first iteration, the initial condition must come from the outside, possibly measured from the actual system.
    } \label{fig:background_PINC_illustration:ab}
\end{figure*}
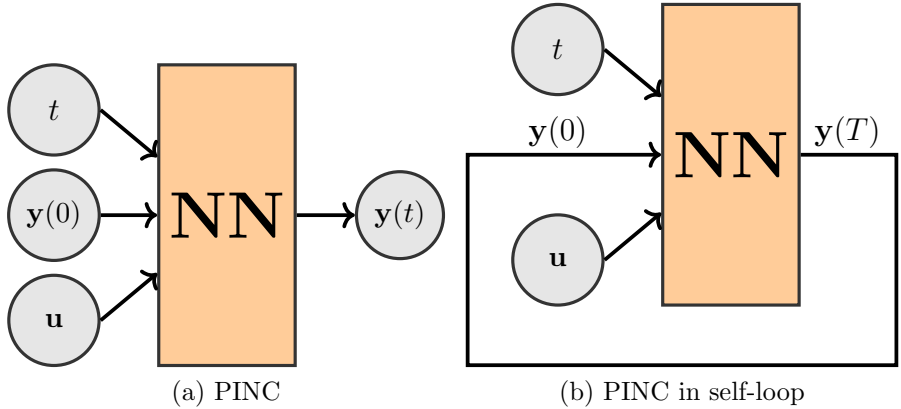

In the traditional PINN framework, the initial condition is fixed and does not support a varying control input. The framework initially proposed in \cite{PINC_paper} adds the initial condition and control signal as inputs to the PINN, leading to the Physics Informed Neural Networks for Control (PINC). 
This argumentation increases the input dimension of the PINN, along with the time required for training, enabling the resulting neural network to make predictions from any initial condition and control input.  
Figure~\ref{fig:background_PINC_illustration} illustrates the concept whereby the PINC maps the input time, the initial condition, and control input to a state prediction at the input time $t$.

Some changes are needed for the training regime to account for the PINC's new inputs (see Figure~\ref{fig:background_PINC_illustration:ab}). Namely, the training data input must be extended with the additional inputs and their ranges selected. 
For the control input $\mathbf{u}$, this should be the feasible range of the control input. 
  The range for the initial condition should contain the states that the system can reach so that the neural network is trained to make predictions from any reachable state during operation.

\subsubsection{Loss Function}
The loss function for PINC is similar to the one for the traditional PINN, \textit{i.e.}, one component for the initial conditions and one for the physics-based loss:
\begin{equation}
    MSE = \lambda_{y} \cdot MSE_{y} + \lambda_{F} \cdot MSE_{F} 
    \label{eq:background_PINC_MSE}
\end{equation}

The only difference here is that we added scaling factors to both terms, which may also be applied to the traditional PINN but was left out for easy understanding. 
    Here, the scaling factors are scalars, but the concept may easily be extended to include individual scaling factors for each of the states so that both $\mathbf{\lambda}_y$ and $\mathbf{\lambda}_F$ can become vectors with the same dimension as $\mathbf{y}$. 
This scaling allows the selection of priority for training of the different states, which may help obtain a better result.

\subsubsection*{Initial Conditions}
For the traditional PINN, only one point was needed to train for the initial condition. Now, however, our initial condition belongs to a range of values, and we also need to consider different control inputs. 
We will create a set of training data points $\{\mathbf{v}^j: j=1,\dots,N_t\}$, where each point contains the inputs to the neural network $\mathbf{v}^j = \left [ 0, \mathbf{y}_0^j, \mathbf{u}^j\right ]^T$. 
The first term is the time $t$, being $0$ for all these data points because they represent the initial condition. 
$\mathbf{y}_0^j$ and $\mathbf{u}^j$ correspond to the $j$-th initial condition and control signal, randomly sampled within their feasible ranges to form $N_t$ data points. 
 The target for each of these training data points is the initial condition, which appears in the input: $\hat{\mathbf{y}}^j= \mathbf{y}_0^j$.   

The loss for the initial conditions is then calculated as the MSE of all these training data points:
\begin{equation}
    MSE_y = \frac{1}{N_t}  \sum_{j=1}^{N_t} \frac{1}{N_y} \left \| \mathbf{y}(\mathbf{v}^j) - \hat{\mathbf{y}}^j \right \|^2 
    \label{eq:background_PINC_MSE_y}
\end{equation}

\subsubsection*{Physics}

The collocation points, $\{\mathbf{v}^k : k=1,\dots,N_f\}$, used to impose the ODE on the PINC net also need to be extended with the new inputs, that is, $\mathbf{v}^k = \left [ t^k, \mathbf{y}_0^k, \mathbf{u}^k\right ]^T$.
 All three elements of each collocation point are sampled randomly from their respective domains.

The residual function needs a slight change to include the control input $\mathbf{u}$ in the ODE function:
\begin{equation}
    \mathbf{F}(\mathbf{y}):= \frac{\partial \mathbf{y}}{\partial t} - \mathbf{f}(\mathbf{y}, \mathbf{u})
    \label{eq:background_PINC_F}
\end{equation}

The physics-related loss term $MSE_F$ is the same as for the traditional PINN, only differing in the extended input to the neural network ($\mathbf{v}^k$):
\begin{equation}
    MSE_F =  \frac{1}{N_f}\sum_{k=1}^{N_f} \frac{1}{N_y} \left \| \mathbf{F}(\mathbf{y}(\mathbf{v}^k)) \right \|^2
    \label{eq:background_PINC_MSE_f}
\end{equation}

\subsubsection{Long Range Simulations by Looping the PINC}

The PINC is trained to make predictions within the time horizon $t \in \left [ 0, T \right ]$, but longer-range simulations are possible by utilizing the PINC several times in a self-loop mode. 
   First, we use the system's current state to make a prediction at time $T$; then, we use this prediction as input to the PINC to obtain a new prediction at time $2T$. 
We can continue this feeding back of the predicted state, simulating $T$ seconds at every iteration until reaching the desired prediction time. 
   Figure~\ref{fig:background_PINC_illustration_self_loop} illustrates how this works, feeding the neural network output back as the initial condition, input to the network, in the next time interval. 

When used with MPC, the prediction time $T$ of the PINC will be selected to match the length of a single step in the controller so that the PINC, with input $t=T$, will act as a function mapping the states between the time steps of the MPC. With this setup, the PINC allows choosing a different control input at every time step, which is perfect for MPC applications.



\subsection{Model Predictive Control} \label{sec:background_MPC}

In this work, Model Predictive Control (MPC) \citep{NMPC_book} will be employed to control an oil well. 
   To ensure that the predictions of the MPC satisfy the system dynamics, we need a function that maps the current state and control input to the state at the next time step. 
For example, for a linear system, the prediction can be implemented by a state transition matrix obtained from the discretization of the system. 
   For a non-linear system, such as the well model, we can use a linearization of the model or a numerical integration method. 
Here, the PINC serves as the function mapping the current state and control input to the state at the next time step. 
  As the PINC contains non-linear functions (the activation functions), the resulting MPC will be within the category called Non-linear MPC (NMPC). 

In this work, the following formulation is considered for model predictive control of an oil well:
\begin{subequations}
\begin{equation}
    \min \sum_{i=1}^{N}(\mathbf{y}[k+i]-\mathbf{y}^{\tt ref})^T \mathbf{Q}(\mathbf{y}[k+i] - \mathbf{y}^{\tt ref}) + \sum_{i=0}^{N_u-1}\Delta \mathbf{u}[k+i]^T \mathbf{R} \Delta \mathbf{u}[k+i]  \label{eq:backgorund_NMPC_y_J}
\end{equation}
  subject to:
  \begin{align}
   & \mathbf{x}[k + j + 1] = \mathbf{F}(\mathbf{x}[k + j], \mathbf{u}[k+j]), ~  j=0,\ldots,N-1 \label{eq:backgorund_NMPC_y_F} \\
    & \mathbf{y}[k+j] = \mathbf{F_y}(\mathbf{x}[k+j], \mathbf{u}[k+j-1]), ~  j=1,\ldots,N  \label{eq:backgorund_NMPC_y_Fy} \\
   & \mathbf{u}[k+j] = \mathbf{u}[k+j-1] + \Delta \mathbf{u}[k+j], ~ j=0,\ldots,N_u-1 \label{eq:backgorund_NMPC_y_u1} \\
   & \mathbf{u}[k+j] = \mathbf{u}[k+N_u-1], ~ j=N_u,\ldots,N-1  \label{eq:backgorund_NMPC_y_u2} \\
     & \mathbf{h}(\mathbf{x}[k+j], \mathbf{y}[k+j], \mathbf{u}[k+j-1]) \leq \mathbf{0}, ~ j=1,\ldots,N  \label{eq:backgorund_NMPC_y_h}
  \end{align}
  \label{eq:backgorund_NMPC_y}
\end{subequations}
where: the prediction horizon of MPC is $N$ steps; the control input is allowed to change for the first $N_u$ steps;
equation \eqref{eq:backgorund_NMPC_y_J} defines the cost function of the optimization problem, whereby the deviations of the output variables from their references and the changes in control input are penalized; 
$\mathbf{Q}$ is the weight matrix for the output deviations from their references;
$\mathbf{R}$ is the weight matrix that penalizes the changes in control inputs for the first $N_u$ iterations;
equation \eqref{eq:backgorund_NMPC_y_F} ensures that the solution satisfies the dynamics of the system, \textit{i.e.}, $\mathbf{F}$ represents the PINC which maps,  one time-step ahead,  states to the next ones; $\mathbf{F_y}$ in equation \eqref{eq:backgorund_NMPC_y_Fy} is a function that maps the state and control inputs to some output variable $\mathbf{y}$, which is the variable we wish to control (the states themselves or algebraic variables, such as the bottom-hole pressure in the oil well).

Equation \eqref{eq:backgorund_NMPC_y_u1} ensures that the first $N_u$ control inputs change by their respective $\Delta \mathbf{u}$ factor. 
While the control input is only allowed to change for the first $N_u$ iterations, afterward, it remains constant and equal to the control input at time step $N_u$ according to equation \eqref{eq:backgorund_NMPC_y_u2}.
Finally, equation \eqref{eq:backgorund_NMPC_y_h} defines the inequality constraints for the controller. They can limit the value of the control inputs $\mathbf{u}$ and also set bounds on states or other variables of interest.

\subsection{Improved PINC}

The improvements proposed for the PINC architecture in this paper, which are presented below, consist of adding skip connections to improve training and avoid vanishing gradients and changing the model's ODE equation to make the training of PINCs with a physics-based loss function feasible.

\subsubsection{Skip Connections}

\begin{figure*}[htb]
  \begin{center}
\begin{tikzpicture}[
rec-node/.style={rectangle},
]
\node[rec-node] (X1) at (0,1)   {$\mathbf{X}$};
\node[rec-node] (X0) at (0,0)   {$\mathbf{X}$};
\node[rec-node] (X-1) at (0,-1) {$\mathbf{X}$};
        
\node[rec-node] (U) at (2.8,1)    {$\mathbf{U}$};
\node[rec-node] (Z1) at (2.8,0)   {$\mathbf{Z}^{(1)}$};
\node[rec-node] (V) at (2.8,-1)   {$\mathbf{V}$};
 
\path[->,line width=0.5mm, red]  (X1) edge node [midway,above] {\small $\mathbf{W}^{1},\mathbf{b}^{1}$} (U);
  \path[->,line width=0.5mm, black]  (X0) edge node [midway,above] 
      {\small $\mathbf{W}^{z,1},\mathbf{b}^{z,1}$} (Z1);
\path[->,line width=0.5mm, red]  (X-1) edge node [midway,above] {\small $\mathbf{W}^{2},\mathbf{b}^{2}$} (V);

\node[rec-node] (A1) at (4.2,0)   {$\mathbf{A}^{(1)}$};
\node[rec-node] (Z2) at (7,0)   {$\mathbf{Z}^{(2)}$};
\path[->,line width=0.5mm, black]  (A1) edge node [midway,above] 
    {\small $\mathbf{W}^{z,2},\mathbf{b}^{z,2}$} (Z2);
\draw[->,line width=0.5mm, black] (Z1) -- (A1);
\draw[->,line width=0.5mm, red,bend right] (U) -- (A1);
\draw[->,line width=0.5mm, red,bend left] (V) -- (A1);

\node[rec-node] (A2) at (8.4,0)   {$\mathbf{A}^{(2)}$};
\draw[->,line width=0.5mm, black] (Z2) -- (A2);
\draw[->,line width=0.5mm, red,bend right] (U) -- (7,1) -- (A2);
\draw[->,line width=0.5mm, red,bend left] (V) -- (7,-1) -- (A2);

\node[rec-node] (AN) at (10.5,0)   {$\mathbf{A}^{(N)}$};
\draw[->,line width=0.5mm, black, dash dot] (A2) -- (AN);
\node[rec-node] (y) at (13.1,0)   {$\mathbf{y}$};
\path[->,line width=0.5mm, black]  (AN) edge node [midway,above] 
    {\small $\mathbf{W}^{z,N},\mathbf{b}^{z,N}$} (y);

\draw[->,line width=0.5mm, red,bend right] (U) -- (8.4,1.4) -- (AN);
\draw[->,line width=0.5mm, red,bend right] (V) -- (8.4,-1.4) -- (AN);

\end{tikzpicture}
\end{center}
       \caption{Illustration of the improved neural network architecture for PINNs, as proposed by \cite{improved_NN_struct}. $\mathbf{X}$ is the input to the neural network, which is projected into a higher dimensional space through the two encoder layers, resulting in the higher dimensional vectors $\mathbf{U}$ and $\mathbf{V}$. For each layer of the main network, we first calculate the intermediate activation $\mathbf{Z}$ utilizing the weight matrix and bias vector of this layer, then $\mathbf{Z}$ is weighted by the encoder outputs $\mathbf{U}$ and $\mathbf{V}$ to calculate the final activation $\mathbf{A}$ of the layer.}
    \label{fig:implementation_improved_NN_structure}
\end{figure*}
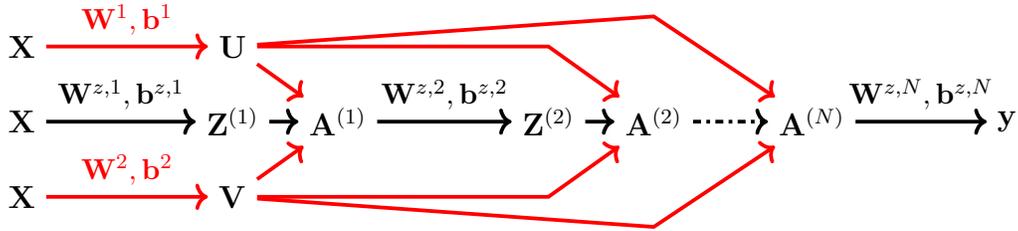

Since gradients can vanish unwittingly during PINC network training, which decreases prediction performance, especially for the particular application of oil well modeling, we propose to employ an improved PINC architecture in this section.
    Any strategy that improves the magnitude of the gradients can greatly help train the neural network. 
One approach is to use skip connections in the network architecture to avoid vanishing gradients, as proposed by \cite{improved_NN_struct}. 
   Besides the fully connected dense network, their proposed structure adds two more layers called encoders.
The input to these encoder layers is the neural network input. In contrast, the output of these encoders is used to calculate the activation of each layer in the fully connected network, except for the final layer. 
   This skip-connection structure ensures that the output of each layer of the neural network is closely connected to the input layer. 
 Equation \eqref{eq:implementatin_improved_NN_structure} expresses the mathematical formulation of the proposed structure,
\begin{subequations}
\begin{align}
   & \left \{ \begin{array}{l} \mathbf{U} =\boldsymbol{\phi}\left( \mathbf{W}^{1}\mathbf{X}+\mathbf{b}^{1}\right) \\
   \mathbf{V}=\boldsymbol{\phi}\left( \mathbf{W}^{2}\mathbf{X}+\mathbf{b}^{2}\right) 
      \end{array} 
      \right .  \label{eq:implementatin_improved_NN_encoder}  \\
    & \left \{ \begin{array}{l} \mathbf{Z}^{(1)} = \boldsymbol{\phi}\left(\mathbf{W}^{z, 1}\mathbf{X}+\mathbf{b}^{z, 1}\right)\\
        \mathbf{A}^{(1)}=\left(1-\mathbf{Z}^{(1)}\right) \odot \mathbf{U}+\mathbf{Z}^{(1)} \odot \mathbf{V}
          \end{array}
          \right . \label{eq:implementatin_improved_NN_Z1_A1}\\
   & \left \{ \begin{array}{l} \mathbf{Z}^{(k)}=\boldsymbol{\phi}\left( \mathbf{W}^{z, k}\mathbf{A}^{(k-1)}+\mathbf{b}^{z, k}\right),  \\
     \mathbf{A}^{(k)}=\left(1-\mathbf{Z}^{(k)}\right) \odot \mathbf{U}+\mathbf{Z}^{(k)} \odot \mathbf{V}, \quad k=2, \ldots, N_L 
        \end{array}
       \right .   \label{eq:implementatin_improved_NN_A_Z}  \\
   & \mathbf{y} = \mathbf{W} \mathbf{A}^{(N_L)}+b \label{eq:implementatin_improved_NN_y}
\end{align}
\label{eq:implementatin_improved_NN_structure}
\end{subequations}
where $\odot$ denotes the Hadamard product, namely the element-wise product of two matrices.
  The visual representation of the structure of skip connections is depicted in Figure~\ref{fig:implementation_improved_NN_structure}.

This structure implements a neural network with $N_L$ hidden layers of $N_n$ neurons each. 
  $\mathbf{X}$ is the input to the network, which, for our application, is a vector containing the time, the initial condition, and the control input. 
The input $\mathbf{X}$ is projected into a higher dimensional space through the encoders layers by using their respective weights and biases $\mathbf{W}^1$, $\mathbf{W}^2$, $\mathbf{b}^1$, and $\mathbf{b}^2$, resulting in the $N_n$ dimensional vectors $\mathbf{U}$ and $\mathbf{V}$.    $\boldsymbol{\phi}$ is the hyperbolic tangent activation function.  

The first step for calculating the forward pass is a standard pass through a dense layer to compute $\mathbf{Z}^{(1)}$, as shown in Equation \eqref{eq:implementatin_improved_NN_Z1_A1}. 
   Then this $\mathbf{Z}^{(1)}$ is weighted by element wise multiplication ($\odot$) with the encoder vectors $\mathbf{U}$ and $\mathbf{V}$, to calculate the activation $\mathbf{A}^{(1)}$ of the first layer. This propagation continues through all the remainder of the $N_L$ layers.  
The final output $\mathbf{y}$ is calculated as a linear projection of the previous layer's activation without the weighting from the encoder layers.

\subsubsection{Safeguarding the Learning Process}

Some of the ODE equations used to calculate $MSE_F$ can lead to problems during training, so some modifications are needed for the neural network to train reliably. Modifying the ODE equations constitutes a problem that depends on the type of system we are interested in modeling with the PINC network. It happens that the oil well's ODEs have undesirable functions that can cause the training process to crash. Two of these problems are addressed in the following.

\textbf{Tubing Friction Factor}:
Equation \eqref{eq:well_lambda_tb} is highly non-linear and not defined for negative numbers inside the logarithm function. For the logarithm argument to be negative, the Reynolds number in the tubing must be sufficiently negative, which is not physically possible. However, such an undesirable condition arises during the initial phase of the training. 
To simplify this equation and help accelerate learning, we replace the equation with an approximation using a third-order polynomial in the region of interest of Reynolds numbers, which can be found through test simulations.

During the initial training phase, the neural network will output poor predictions that can be far from physically feasible values. We can then end up with Reynolds numbers outside the interval for which the approximation was fit. The most straightforward tweak is to truncate the Reynolds number (calculated from the PINC prediction) to the selected interval. Because this truncation eliminates the dependency on the preceding variables, it also reduces the quality of the gradients when the Reynolds number gets truncated, but, in practice, the truncation works just fine. Equation \eqref{eq:model_approx_Re_tb} formalizes the approximation resulting from the truncation of the Reynolds number, with $g(\cdot)$ being the third-order polynomial approximation defined in Equation \eqref{eq:model_approx_Re_tb_third_order_function}.
    \begin{subequations}
\begin{align}
    \lambda_{\tt tb}(Re_{\tt tb})&=\left\{\begin{matrix}
g(Re_{{\tt tb},\min}), & Re_{\tt tb} < Re_{\tt tb,\min}\\
g(Re_{\tt tb}), & Re_{{\tt tb},\min} \leq Re_{\tt tb} \leq Re_{{\tt tb},\max}\\ 
g(Re_{{\tt tb},\max}),  & Re_{\tt tb} > Re_{{\tt tb},\max}
\end{matrix}\right. 
\label{eq:model_approx_Re_tb}  \\
    g(x) &= ax^3 + bx^2 + cx + d
\label{eq:model_approx_Re_tb_third_order_function}
\end{align}
\end{subequations}

The numerical values for the polynomial coefficients and the ranges of Reynolds numbers used to generate these approximations are shown in Table~\ref{tab:implementation_3rd_order_approx_coeficients}. 
\ref{app:odechange} provides more details on the implemented approximation.
The oil well model also contains a friction factor for the bottom-hole friction, utilizing the same equations. However, since the flow from the reservoir is constant when calculating the Reynolds number here, this friction factor is also constant and no changes are needed.

\textbf{Square Root:}
The square root is another function that can cause numerical problems, mainly during the initial phase of the training, because it is not defined for negative numbers. If the neural network outputs state predictions that result in a negative value in one of the square roots in the ODE, the resulting value will not be defined (``NaN'' in Python). This ``NaN'' (Not a Number) will propagate through the calculations, invalidating the whole training.

The issue is resolved by simply applying the $\max(\cdot)$ operator within these functions to avoid negative values. Even though this tweak results in no gradients for negative numbers, it worked well since this mainly occurs in the initial iterations of the training.  
To avoid zero division error in other equations, it is helpful to set the lower limit of the square root input to some small positive value (\textit{e.g.}, $10^{-3}$), a threshold that can be found by trial and error.
This modified square root function is:
\begin{equation}
    f_{\tt sqrt}(x) = \sqrt{\max \left \{x,  10^{-3}\right \}}
    \label{eq:implementation_sqrt_w_max}
\end{equation}

\subsection{Neural Network for Algebraic Variables}
\label{sec:alg_NN}

As we are interested in controlling the algebraic states of the oil well model, for instance, the bottom-hole pressure or liquid production, we need some means for obtaining the values of these variables as the current PINC framework only outputs the state predictions.

\begin{figure*}[htb!]
\centering
\begin{tikzpicture}[
orangenode/.style={rectangle, draw=black!80, fill=orange!40, very thick, minimum size=7mm},
greennode/.style={circle, draw=black!80, fill=green!30, very thick, minimum size=5mm},
graynode/.style={circle, draw=black!80, fill=black!10, very thick, minimum size=5mm},
]
\node[orangenode,minimum height=4cm]      (nn) at (0,0)   
         {\Huge \textbf{PINC} };
\node[graynode,minimum size=1.2cm] (t) at (-3.2,1.4) {\small $t$}; 
\draw[->,line width=0.5mm, black] (t.east) -- (-1.4,1.4);
  \node[graynode,minimum size=1.2cm] (y0) at (-3.2,0) {\small $\mathbf{y}(0)$}; 
   \draw[->]   [line width=0.5mm, black] (y0) -- (nn);
\node[graynode,minimum size=1.2cm] (u) at (-3.2,-1.4) {\small $\mathbf{u}$}; 
\draw[->,line width=0.5mm, black] (u.east) -- (-1.4,-1.4);

\node[orangenode,minimum height=2cm]      (nn-y) at (4.2,0)   
         {\Huge \textbf{NN} };
\path[->,line width=0.5mm, black] (1.4,0.3) edge node [midway,above] 
    {\small $\mathbf{y}(T)$}  (3.3,0.3); 
    
\draw[->,line width=0.5mm, black]  (-2.1,-1.4) -- (-2.1,-2.8) -- (2.5,-2.8)
       -- (2.5,-0.5) -- (3.3,-0.5); 
         
\node[graynode,minimum size=1.2cm] (zT) at (7,0) {\small $\mathbf{z}(T)$};         
     \draw[->]   [line width=0.5mm, black] (nn-y) -- (zT);
     
\end{tikzpicture}
 \caption{Hierarchical architecture for computing algebraic variables from the states.
    An additional, independently trained neural network is trained to predict the algebraic variables of the oil well, for instance, the bottom-hole pressure. The inputs to this supporting network are the state predictions from the PINC and the control input.}\label{fig:implementation_PINC_with_alg_NN}

\end{figure*}
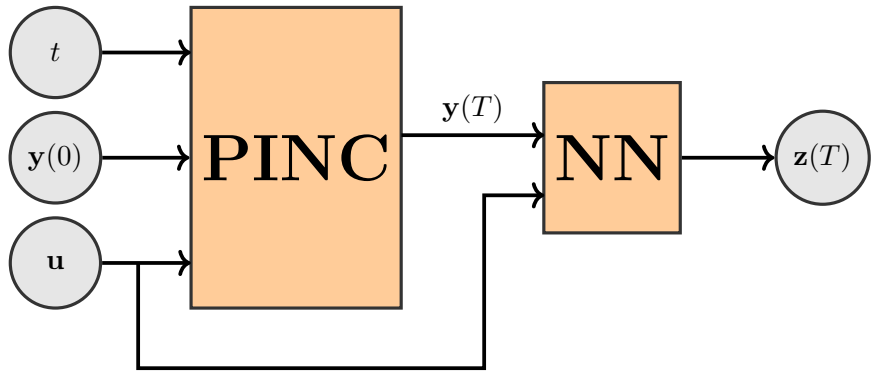

Although PINC could be trained to output the algebraic variables directly, this would limit its flexibility, as additional algebraic variables would require fully retraining the network from scratch.
   Thus, we propose to train an independent, traditional feedforward neural network
to predict the algebraic variables $\mathbf{z}$ from the states $\mathbf{y}$. 
   The input to this neural network are the states $\mathbf{y}$ and the control input $\mathbf{u}$, while the output is one or more of the algebraic variables $\mathbf{z}$ of the model. This setup is illustrated in Figure~\ref{fig:implementation_PINC_with_alg_NN}. 
This new neural network is trained independently of the PINC. 
The hierarchical architecture allows
utilizing different algebraic variables in the MPC by simply training a new neural network without retraining the PINC model, which is much slower and more complex.

As no dynamics are involved in calculating the algebraic variables from the states, the PINN framework is unnecessary. Like the PINC, training data is sampled from the entire state and control-input space. The corresponding target outputs (the values of the algebraic variables) are calculated using the ODE equations. 
   \bb{It is important to note that modeling the entire system in an end-to-end manner with neural networks, \textit{i.e.},  mapping inputs to the desired algebraic variables, allows the MPC implemented in CasADi\footnote{CasADi is an open-source tool for nonlinear optimization and algorithmic differentiation, which can be downloaded from \url{https://web.casadi.org/}} to leverage a smooth and differentiable function (\textit{i.e.}, the neural network itself) as the system's model. 
       Thus, using a network model enables the computation of a Jacobian without hindrances.
  CasADi solves the resulting NLP problem with IPOPT \citep{IPOPT2006}, an interior-point algorithm developed for problems with smooth and twice differentiable objective functions and constraints. IPOPT is a robust NLP solver widely applied in NMPC and optimal control.}

\bb{Empirical observations revealed that relying solely on the ODE equations to compute algebraic variables from the PINC's state prediction presents challenges for the MPC in CasADi.
In practice, this approach proved ineffective, as the optimization algorithm in CasADi often crashed before completing a meaningful simulation, with no discernible pattern to the crashes. It is likely that the nature of the equations, involving square roots, logarithms, and maximum operators, occasionally caused the optimizer to crash. This observation further underscores the rationale behind adopting the proposed hierarchical architecture.}

Particularly for our oil well application,
\bb{this supporting neural network will be trained to predict four algebraic variables based on the input from PINC's state predictions and control input}, \textit{i.e.}, the desired output $\mathbf{z}$ is a four-dimensional vector: 
\begin{itemize}
    \item \textit{bottom-hole pressure} ($P_{\tt bh}$): the subject of control for one MPC application.
    
    \item \textit{Gas-lift injection rate} ($w_{{\tt G},{\tt in}}$): a limited resource the controller needs to take into account.
    
    \item \textit{Gas production} ($w_{{\tt G},{\tt res}}$): it may need to be constrained as some plants have limited capacity for processing gas.
    
    \item \textit{Liquid production} ($w_{{\tt L},{\tt res}}$):  it  can be used for economic optimization.
\end{itemize}

\subsection{PINC for the Oil Well}
\bb{Figure~\ref{fig:implementation_PINC_with_alg_NN} depicts the architecture chosen for the PINC, where the PINC module employs the structure with skip connections. This PINC module has three types of inputs: time $t$; 
two control inputs corresponding to the production choke opening $u_1$ in \eqref{eq:well_w_out} and the gas-lift valve opening $u_2$ in \eqref{eq:well_w_G_in};
and an initial condition corresponding to a three-dimensional vector with the mass of gas in the annulus $m_{{\tt G},{\tt an}}$, mass of gas in the tubing $m_{{\tt G},{\tt tb}}$, and mass of liquid in the tubing $m_{{\tt L},{\tt tb}}$. The physics-informed loss function corresponds to Eq. \eqref{eq:well_ode} and equations from \ref{app:well}.
In total, the input layer has six units, 
while the output layer has three units mapping the states
 $m_{{\tt G},{\tt an}}$, $m_{{\tt G},{\tt tb}}$, and $m_{{\tt L},{\tt tb}}$ at time $t$.
 }

\bb{The NN module has a three-dimensional input layer, which receives the output prediction of the PINC module of the same dimension, \textit{i.e.}, the predicted states
$m_{{\tt G},{\tt an}}$, $m_{{\tt G},{\tt tb}}$, and $m_{{\tt L},{\tt tb}}$ at time $t$.
It outputs a four-dimensional vector consisting of the algebraic variables described in the previous section, namely 
$P_{\tt bh}$, 
$w_{{\tt G},{\tt in}}$,
$w_{{\tt G},{\tt res}}$, and
$w_{{\tt L},{\tt res}}$. 
The bottom-hole pressure $P_{\tt bh}$ will be subject to control in our experiments by manipulating the two control inputs (valve openings).
}


\subsection{Evaluation Metrics}

\subsubsection{Model Validation}
During hyperparameter search, the trained PINC is evaluated on a validation set to select the hyperparameter values that yield the best network performance on unseen data. 
Thus, the MSE between the PINC output and the Runge-Kutta (RK) simulation of the gas-lifted oil well, which represents the actual system, defines the validation loss, namely:
\begin{equation}
    MSE_{\tt val} = \frac{1}{N_{\tt val}N_y}  \sum_{i=1}^{N_{\tt val}} \left \| \mathbf{N}(T, \mathbf{y}_{0,{\tt val}}^i, \mathbf{u}_{\tt val}^i) - \hat{\mathbf{y}}_{\tt val}^i \right \| ^2
\end{equation}
where: 
$N_y$ is the number of states of the model;
     $N_{\tt val}=100$ is the number of validation points drawn randomly considering the chosen ranges for the initial conditions $\mathbf{y}_{0,\tt val}^i$ and control inputs $\mathbf{u}_{\tt val}^i$. 
The target points $\hat{\mathbf{y}}_{\tt val}^i$ are feasible points obtained from RK simulation;
   $\mathbf{N}(T, \mathbf{y}_{0,{\tt val}}^i, \mathbf{u}_{\tt val}^i)$ is the PINC network prediction $T=60$ seconds ahead in time for validation point $i$; and
$\left \| \cdot \right \|$ is the $\ell_2$-norm.

Due to the issue of running into negative values in square roots, the self-looping simulation was not performed for model evaluation as proposed in \citep{PINC_paper}, since these simulations would quickly run into the infeasible state for our application. 
   We assume that if PINC can yield good predictions from any initial condition (within the selected domain), it should also perform well in self-loop. 

\subsubsection{Evaluation of Prediction and Control}

The evaluation of PINC predictions consists of iterating the network in self-loop mode, starting from random initial conditions and with a random sequence of control inputs, and then comparing its prediction to a Runge-Kutta simulation of the system. As we consider the ODE to represent the system perfectly, the Runge-Kutta simulation will represent the actual behavior of the oil well. The performance will be evaluated both visually and using the Integral of Absolute Error (IAE) on $C$ sampling points along the simulation:
\begin{equation}
    IAE = \frac{1}{C} \sum_{k=1}^{C} \frac{1}{N_y} \left \| {\mathbf{y}[k] - \mathbf{r}[k] }\right \|,
    \label{eq:problem_IAE}
\end{equation}
where $\mathbf{y}[k]$ is the output of the PINC network at the sample point $k$ and $\mathbf{r}[k]$ is the Runge-Kutta reference simulation at the same sample point.


\section{Experimental Setup} \label{sec:experiments}

An improved PINC with skip connections, as detailed in the previous section, will be selected as a model of a gas-lifted oil well.  Then, the control of the bottom-hole pressure of the oil well is tackled by embedding the trained PINC in a nonlinear MPC task.

This section presents the techniques and configurations implemented for training a PINC network with the best hyperparameters and the setup for applying nonlinear MPC using the trained PINC network. 

\subsection{Experimental Setup of PINC}

\subsubsection{Domain of Training Data}
\label{sec:domain_of_tarining_Data}

The training data and collocation points are generated as presented in Section \ref{sec:pinc_loss}. In addition,
the upper limit for the time input $t$ should be the prediction time of interest. For our application, the PINC prediction time is the time of a single step of the MPC, \textit{i.e.}, $T=60$ seconds. 
For the control input, the range corresponds to the feasible valve openings \bb{(production choke and gas-lift valve openings)}. 
The range for the states was selected by running several Runge-Kutta simulations with different control inputs to determine the reachable states during operation.
For the oil well model, the resulting domain of these data is a cube in the three-dimensional state space ($m_{{\tt G},{\tt an}}$, $m_{{\tt G},{\tt tb}}$, and $m_{{\tt L},{\tt tb}}$), which may be suboptimal, as some volumes within this cube may be infeasible (\textit{e.g.}, square root of a negative number). 
Infeasible points in the randomly drawn points from this cube are removed by running them through the ODE to check their feasibility. Excluding these infeasible points from the cube yielded the final training set for the states.
Removing these infeasible points was very important, especially for the second well model.

For the initial condition (state), the procedure above is directly applied to generate the respective training set.
For the collocation points, however, this procedure only verifies that the initial condition is feasible, not that the desired output prediction 60 seconds ahead of time exists. 
So, if the initial condition is close to infeasibility and the control input is unfortunate, the desired output may not be feasible. In practice, this works fine, but improvements may be possible.

\subsubsection{Hyperparameters Search}

During the hyperparameter search, all tentative configurations were trained $10$ times with distinct initializations of the network weights and biases. 
For performance evaluation, the error on the validation set was considered.

\subsubsection*{Adam Learning Rate}

For this search, we trained a PINC net with 5 hidden layers consisting of 20 neurons each for 1,000 iterations using the following learning rates: $\{1,4,7\}\times 10^{-4}$, $\{1,4,7\}\times 10^{-3}$, and $\{1,4\}\times 10^{-2}$, resulting in the best choice of $7\times 10^{-3}$, \bb{which presented lowest validation prediction error}.



\subsubsection*{Number of Training Data Points and Collocation Points}
\begin{figure*}[tb!]
    \centering
    \includesvg[width=0.7\textwidth]{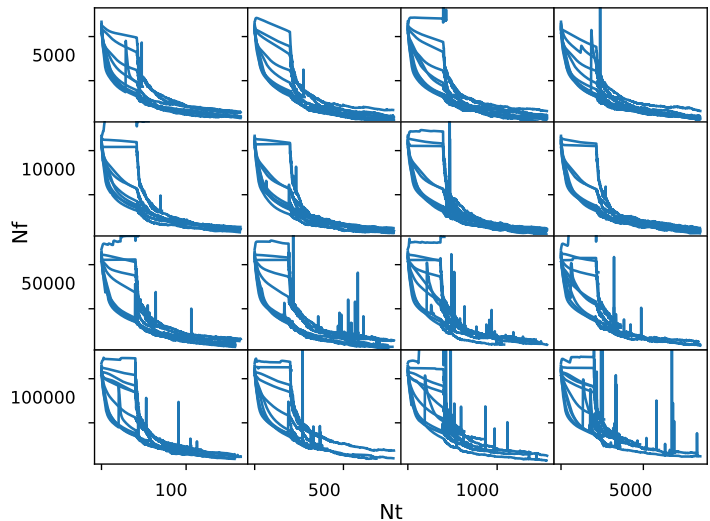}
    \caption{Validation error progression during PINC training with different combinations for the values of $N_t$ (no. of data points) and $N_f$ (no. of collocation points). 
    In each subplot, the lines represent the training evolution of ten randomly initialized neural networks. 
    The limits on the y-axis are the same for all the subplots. The best average performance was attained with \mbox{$N_t$=1,000} and \mbox{$N_f$=10,000}, wherein none of the training attempts terminated prematurely.
    }
    \label{fig:implementation_hp_search_Nt_Nf}
\end{figure*}

To find the number $N_t$ of data points for the initial condition  and the number $N_f$ of collocation points, the same network of 5 hidden layers of 20 neurons each was employed. 
For each training run, after \mbox{1,000} epochs of Adam\footnote{
\bb{On average, using the Adam method in the first thousand iterations yielded lower prediction validation error and faster training than using only L-BFGS optimization through the whole training, although both errors were in the same order of magnitude.}}, 
other \mbox{1,000} iterations of L-BFGS followed. 
Figure~\ref{fig:implementation_hp_search_Nt_Nf} shows plots of the validation error during training of ten models for each combination of $N_t$ and $N_f$. 
The y-axis has the same limits in all the plots. Some training experiments terminate prematurely due to converging to a poor local minimum. Considering both the final validation error and the number of training trials that fail, the best choice was the combination \mbox{$N_t$=1,000} and \mbox{$N_f$=10,000}. 

\subsubsection*{PINC Architecture Search}
\begin{figure*}[tb!]
    \centering
    \includesvg[width=0.7\textwidth]{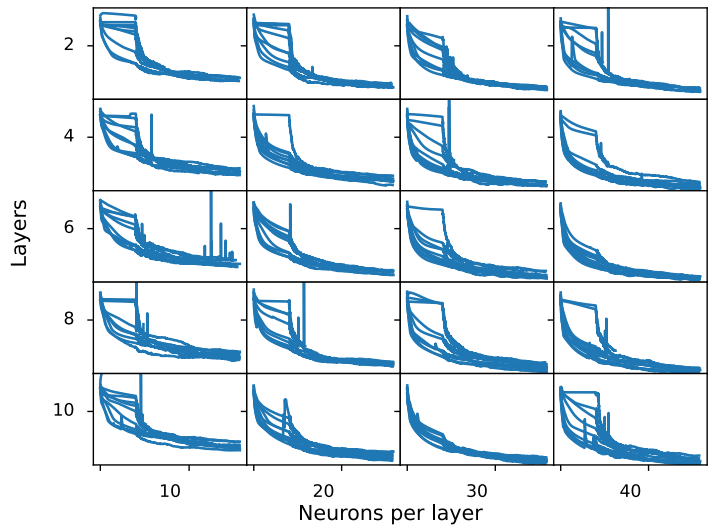}
    \caption{Search for the PINC architecture, presenting the validation error during the training of ten neural networks for each configuration, consisting of the number of hidden layers and neurons per layer. 
    The best three networks are $6\cdot 30$, $6\cdot40$, and $10\cdot30$ layers and neurons per layer.
    The y-axis has the same limits in all the subplots.}
    \label{fig:implementation_hp_search_NN_struct_1}
\end{figure*}

To find the right number of hidden layers and neurons per layer, a grid search was executed for $\{2,4,6,8,10\}$ layers and $\{10,20,30,40\}$ neurons per layer, where each network was trained for  \mbox{1,000} epochs with Adam followed by \mbox{1,000} iterations with L-BFGS (Figure~\ref{fig:implementation_hp_search_NN_struct_1}).
In a second step, the best three networks ($6\cdot 30$, $6\cdot40$ and $10\cdot30$ layers and neurons) were trained for additional \mbox{1,000} iterations with L-BFGS, yielding the architecture of $6$ hidden layers of $30$ neurons each with the lowest average validation loss.

\subsection{Experimental Setup of the NN for Algebraic Variables}

We performed a hyperparameter search to find the best NN architecture for predicting the algebraic variables (the supporting neural network from Figure~\ref{fig:implementation_PINC_with_alg_NN}).   
The structures tested were fully densely connected neural networks with $1$-$4$ hidden layers, each containing $10$, $20$, $30$, and $40$ neurons. Ten tests were conducted for each configuration, with different weight initializations, and trained for \mbox{5,000} iterations of L-BFGS. 
The validation set was constructed by randomly sampling \mbox{10,000} points from the input space, where the respective target values resulted from solving the oil-well ODE equations.
The network of $4$ hidden layers of $30$ neurons each achieved the lowest validation error and was subsequently trained until convergence with L-BFGS. 
However, to speed up the MPC, a smaller network could be employed with no significant decrease in performance.

\subsection{MPC Setup}
The MPC controller was implemented in CasADi, an open-source numerical optimization framework\footnote{\url{https://web.casadi.org/docs/\#document-ocp}}. 

The prediction horizon $N$ of MPC depends on the control problem, which was chosen so that the system reaches a steady state within the specified prediction horizon.
The time step of the predictions is $T=60$ seconds for all control applications.

In typical applications, the optimization system defines a steady state, maximizing total oil production while adhering to system and physical constraints \citep{RauhTOP:2022}. The optimization system informs the resulting bottom-hole pressure to the control system as a reference for tracking. This pressure reaches its reference quickly and usually before the states stabilize. 
    This fast reference tracking happens because of infinitely many combinations of state values and control inputs that will keep the bottom-hole pressure at the reference. 
Thus, a prediction horizon of $N=50$ minutes is needed in this case.
If the states run too far off, the system may end up in a region of the state space where the model is not defined, or the MPC cannot find a solution to the optimization problem. To ensure that a steady state is reached at the end of the prediction horizon, the control-input horizon is selected shorter ($N_u=45$) than the prediction horizon ($N_u<N$). In practice, this means that the control input will be constant for the last 5 steps of the prediction horizon. 
   Also, through the weight matrix $\mathbf{Q}$, an extra penalty is added to these final reference deviations (last 5 steps), making the MPC prioritize finding a stable state rather than damping the transient response:
\begin{equation}
    Q_{ii}=\left\{\begin{matrix}
1, \quad i \in \{1,\dots,45\}  \\  
100,  \quad i \in \{46, \dots, 50\}
\end{matrix}\right. ,\quad
R=\begin{bmatrix}
10^3 & 0\\ 
  0 & 10^3
\end{bmatrix}
\label{eq:methode_MPC_Q_R_BH1}
\end{equation}

\ref{app:mpc} gives more details on the MPC implementation in CasADi.

\section{Results}
\label{sec:results}

Three variations of the oil well model, with different values of parameters as detailed in \ref{app:well}, will be considered, but more results will be shown only with the first model for brevity. The experiments below refer to the first well model if not explicitly stated otherwise.
Both the long-range prediction and nonlinear MPC with the improved PINC for a gas-lifted oil well will be presented in this section.

\subsection{Long-range Prediction for an Oil Well with the Improved PINC}

\begin{figure}[!tb]
    \centering
    \includesvg[width=0.7\textwidth]{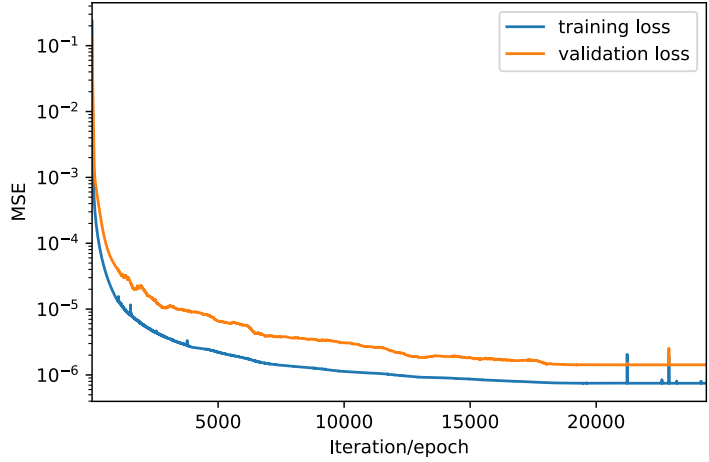}
    \caption{Plot of the training loss and validation error of the PINC trained for prediction of well 1. The PINC was trained for \mbox{1,000} epochs with Adam and then until convergence with L-BFGS. Because the nature of computation for the training loss (composed of the initial condition and physics-based loss) and validation error are distinct, the numeric values of these two losses are not directly comparable. The final training loss is $7.48\cdot10^{-7}$ and the validation loss, $1.43\cdot10^{-6}$.}
    \label{fig:result_PINC_well_1_training}
\end{figure}

The PINC used for prediction and control has 6 hidden layers of 30 neurons each, as found previously in the hyperparameter search. For the first well, Figure~\ref{fig:result_PINC_well_1_training} reports the training loss and validation error during the training of its PINC. The validation loss shown in this figure consists of the average (scaled) error over 100 predictions 60 seconds ahead of time. 

\begin{figure*}[tb!]
    \centering
    \includesvg[width=0.85\textwidth]{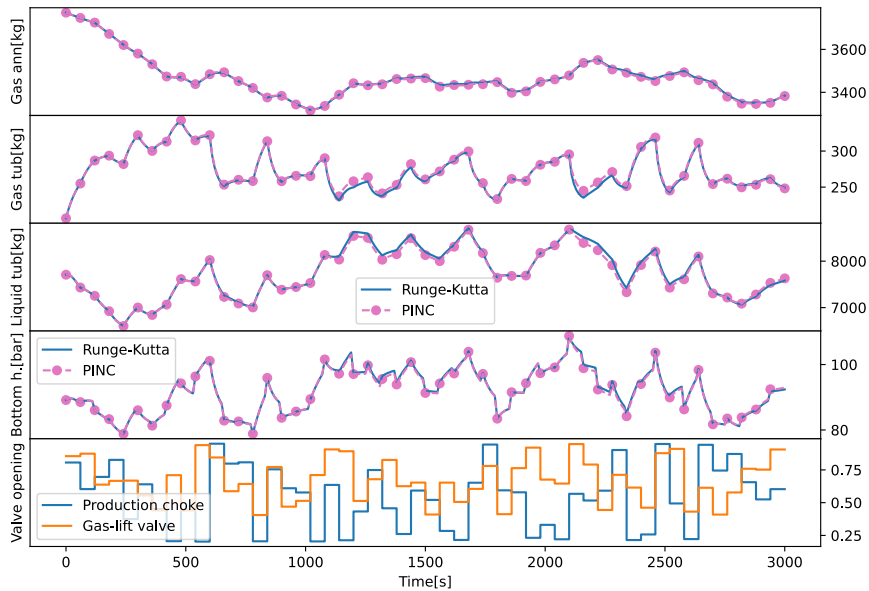}
    \caption{A test simulation of long-range PINC prediction for 3,000 seconds for the first well. A random sequence of control input, changing every 60 seconds, and a fixed (randomly chosen) initial condition are fed to the PINC. The resulting predictions are shown in pink, with the dots indicating each self-loop iteration. The blue line (in the upper three plots) shows the actual state values obtained by the Runge-Kutta simulation. 
    The fourth plot shows the bottom-hole pressure predictions, generated by passing the PINCs predictions through the supporting neural network trained to predict the algebraic states.
    The bottom plot depicts the randomly generated control input sequence. The IAE for the states is 0.00959, a bit below the average, as shown in Table~\ref{tab:result_test_sim_IAE}, while the IAE for the pressure is 0.2813 bar.}
    \label{fig:result_well1_test_sim_1}
    \label{fig:result_well1_test_sim_1_w_Pbh}
\end{figure*}

The performance of the trained PINC was evaluated for \mbox{3,000} seconds, with a sequence of random control inputs that changed every 60 seconds, using the IAE as a performance measure. The PINC was put in a self-loop for 50 iterations to obtain this simulation. 
Only the initial condition and the sequence of control inputs were input to the PINC network so that prediction errors would accumulate over the self-looping. Figure~\ref{fig:result_well1_test_sim_1} shows the long-range PINC prediction for the first well, where the blue line corresponds to the Runge-Kutta simulation (represents the actual system's behavior), the dashed pink line gives the PINC prediction, and the larger pink dots indicate the PINC output after each iteration of self-loop. 
While the first three uppermost plots report the state predictions produced by the PINC, the fourth one shows the bottom-hole pressure predictions generated by feeding the PINCs predictions through the supporting neural network, which predicts the algebraic states.
The bottom-most plot depicts the randomly generated control-input sequence.

Table~\ref{tab:result_test_sim_IAE} shows the average and standard deviation of the IAE (of scaled variables) for 100 test simulations of long-range prediction (3,000 seconds each), where each test simulation resembles the one from Figure~\ref{fig:result_well1_test_sim_1_w_Pbh}, but with different randomly generated control inputs.
  For the first well, the average and highest IAE of all three states are 0.00994 and 0.0332 bar, respectively.
Additionally, the average IAE of the bottom-hole pressure for the first well is 0.3087 bar.  

\begin{table}
  \centering
    \caption{Average and standard deviation of the IAE of the prediction for all three states, and average IAE of the bottom-hole pressure prediction, for three oil wells for 100 test simulations of 3,000 seconds each, with randomly generated inputs. Only the pressure uses the second network for algebraic variables.}
  \label{tab:result_test_sim_IAE}
  
  \begin{tabular}{c|ccc}
  \hline
     IAE & Well 1 & Well 2 & Well 3\\ \hline
     average (states) & 0.00994 & 0.00448 & 0.00185\\
     standard deviation (states) & 0.00621 &  0.00345 & 0.000325\\
     average (pressure) & 0.3087 &  0.2497 & 0.0656\\
  \end{tabular}

\end{table}


\subsection{Nonlinear MPC of Oil Well with Improved PINC}
\begin{figure*}[tb!]
    \centering
    \includesvg[width=0.85\textwidth]{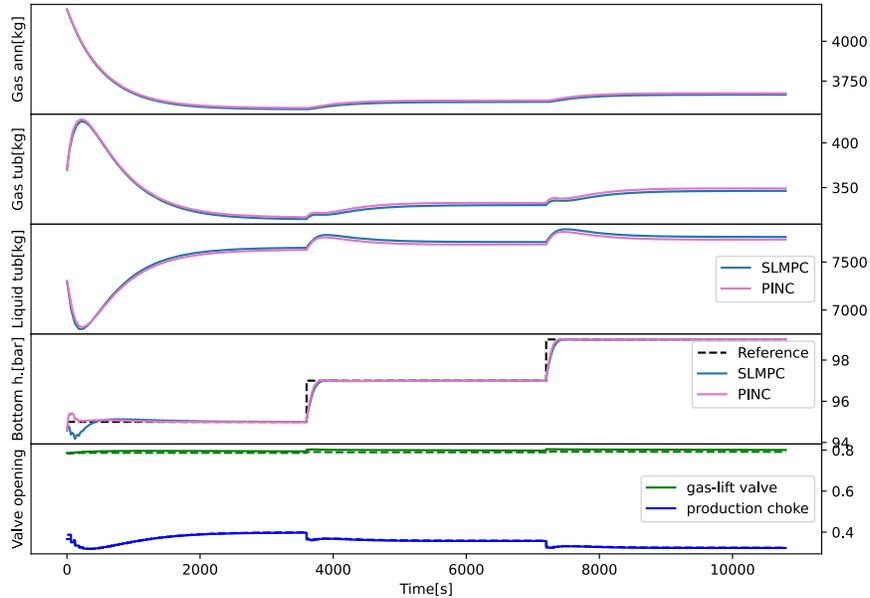}
    \caption{Controlling the bottom-hole pressure ($P_{\tt bh}$): simulation of three-step changes of 2 bar to the bottom-hole pressure reference (95 to 99 bar) for the PINC MPC in pink and the SLMPC in blue. 
    For the control input in the bottom-most plot, the PINC MPC control input is shown in solid lines, and the SLMPC in dashed lines. The upper three plots show the states that are not controlled or constrained ($m_{{\tt G, an}}$, $m_{{\tt G, tb}}$, and $m_{{\tt L, tb}}$). The IAE of the PINC and SLMPC are 0.038 [bar] and 0.058 [bar], respectively.}
    \label{fig:result_MPC_Pbh_sim1}
\end{figure*}

Here, experiments on tracking the bottom-hole pressure with the improved PINC and nonlinear MPC will be shown and compared to a linear MPC of a model linearized at each timestep, namely SLMPC (\ref{app:SLMPC}).
Figure~\ref{fig:result_MPC_Pbh_sim1} shows a simulation with two-step changes in the bottom-hole pressure reference. 
The states are shown in the figure to demonstrate that they settle after some time, even though they are not controlled or constrained in any way ($m_{{\tt G, an}}$, $m_{{\tt G, tb}}$, and $m_{{\tt L, tb}}$). 
 The bottom-hole pressure follows the given reference very well for the PINC MPC controller, with an IAE of 0.038 bar. For comparison, SLMPC, a method that linearizes the ODE equation at every iteration of the control loop, described in \ref{app:SLMPC}, is also plotted in blue, with an IAE of 0.058 bar.
Notice that the biggest challenge in this experiment corresponds to the initial timesteps, where the bottom-hole pressure deviates from the reference for both controllers.

\begin{figure*}[tb!]
    \centering
    \includesvg[width=0.85\textwidth]{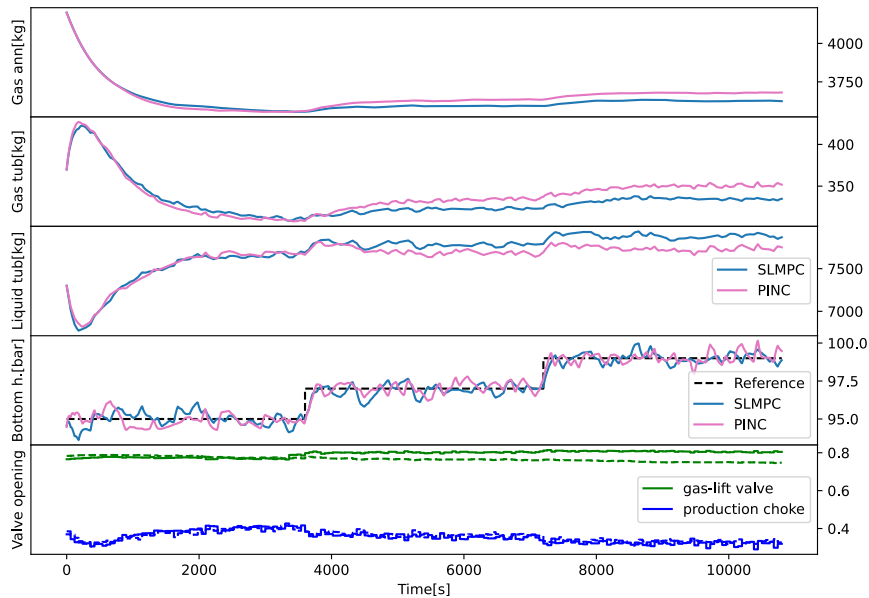}
    \caption{
    Controlling the bottom-hole pressure  ($P_{\tt bh}$) with the addition of measurement noise. In the bottom plot, the PINC MPC control input is shown in solid lines, and the linearized MPC is shown in dashed lines. The IAE of the PINC and SLMPC are 0.317 [bar] and 0.327 [bar], respectively.}
    \label{fig:result_MPC_Pbh_sim1_noise_ex}
\end{figure*}

\begin{figure}[tb!]
    \centering
    \includesvg[width=0.85\textwidth]{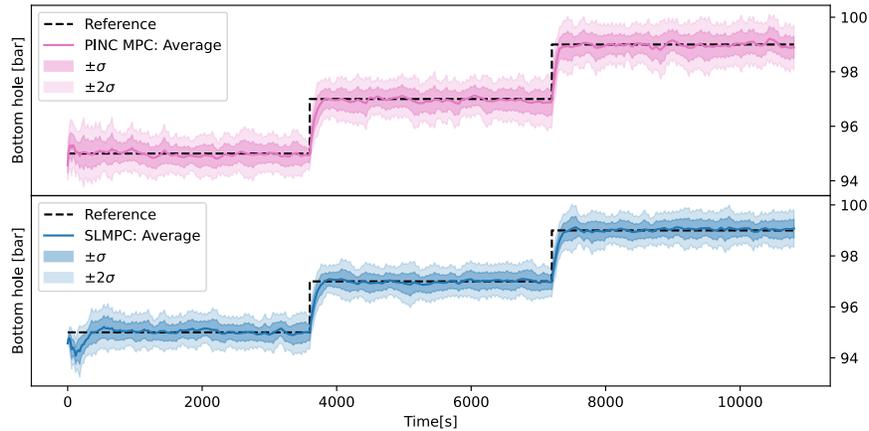}
    \caption{Controlling the bottom-hole pressure  ($P_{\tt bh}$) in several simulations with measurement noise: the average response and the standard deviation are plotted for both controllers. The upper (lower) plot shows the results of the PINC MPC (SLMPC) controller, where the innermost shaded area includes one standard deviation of the response (around 68\% of the total), and the outer shaded area corresponds to two standard deviations of the response (around 95\% of the total).
    }
    \label{fig:result_MPC_Pbh_sim1_PINC_avg_std}
\end{figure}

The same simulation was repeated several times with the addition of white measurement noise with a standard deviation of $5\%$ added to the state measurements. Figure~\ref{fig:result_MPC_Pbh_sim1_noise_ex} shows an example of these simulations.
Several of these simulations were completed, and the IAE was computed for each. The resulting average and standard deviation statistics are shown in Table~\ref{tab:result_Pbh_sim1_nois_IAE}, along with the IAE of the noiseless experiment.
\begin{table}
  \centering
    \caption{IAE of simulations with the two MPCs. The $0\%$ noise experiment is the one shown in Figure~\ref{fig:result_MPC_Pbh_sim1},  while the $5\%$ noise statistics are calculated from several simulations, one of which is shown in Figure~\ref{fig:result_MPC_Pbh_sim1_noise_ex}.}
  \label{tab:result_Pbh_sim1_nois_IAE}
  
  \begin{tabular}{cc|cc}
  \hline
    Noise &  & PINC MPC & SLMPC \\ \hline
    $0\%$ & IAE & 0.038 & 0.058 \\ \hline
     \multirow{2}{*}{$5\%$}& IAE average & 0.339 & 0.297 \\
     & IAE standard deviation & 0.0200 &  0.0207\\
  \end{tabular}

\end{table}
Regarding the simulations with measurement noise, Figure~\ref{fig:result_MPC_Pbh_sim1_PINC_avg_std} reports the average response and the standard deviation, calculated along the trajectory. This figure shows in the solid line the average response, and the two shaded areas show one and two standard deviation regions, which encompass approximately $68\%$ and $95\%$ of the simulations, respectively. 

\subsection{Ablation of the Skip Connections}
\begin{figure}[tb!]
    \centering
    \includesvg[width=0.9\textwidth]{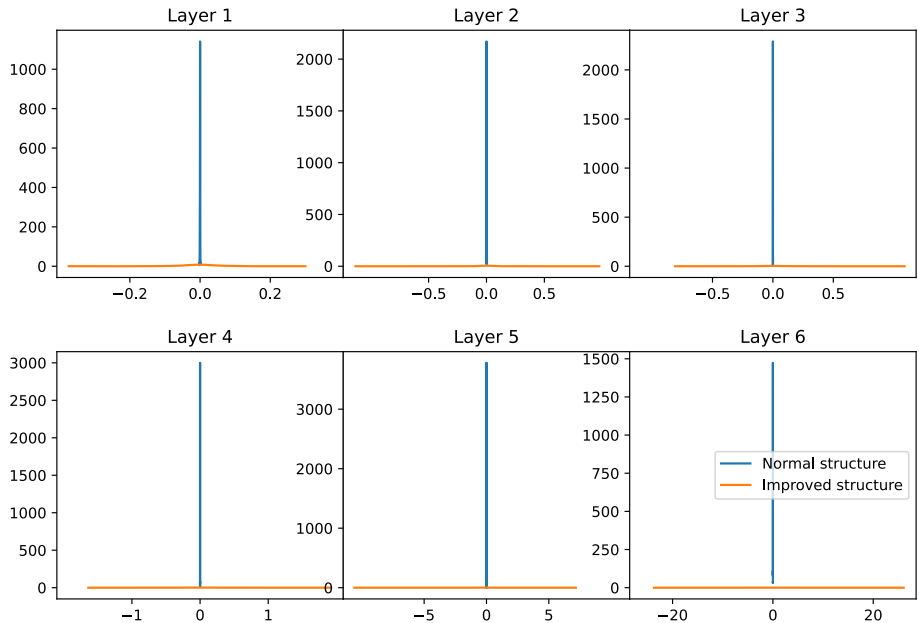}
    \caption{Kernel Density Estimate (KDE) plot comparing the gradients concerning the weights and biases of the residual function 
    $\mathbf{F}(\cdot)$ defined in (9).
    The difference between the two PINC architectures is so expressive that the distributions of gradients for the normal structure (blue) seem like vertical lines compared to the orange horizontal lines for the improved structure (orange, with skip connections). 
    For the \bb{last hidden} layer \bb{(Layer 6)}, the average magnitude of the gradients for the normal structure is $2.16\cdot10^{-4}$, while for the improved structure, it is $2.55$ \bb{(increasing by four orders of magnitude)}.} 
\label{fig:implementation_NN_gradient_decomposition_normal_and_improved_2nd_term}
\end{figure}

In this section, we investigate how skip connections affect the training of PINC. Figure~\ref{fig:implementation_NN_gradient_decomposition_normal_and_improved_2nd_term} shows the Kernel Density Estimate (KDE) plot for the gradients of the residual $\mathbf{F}(\cdot)$ defined in \eqref{eq:background_PINC_F}, for two PINCs with six layers that are partially trained, one with skip connections (improved structure) and another without them (normal structure). The plots show the distribution of the gradients centered around zero for the weights and biases in all the layers. A high spike at the curve's center means that many of the gradients are near zero, which is undesirable. 
The orange line is flat whereas the blue line is vertical, showing that the gradients obtained with the improved structure are much better than the gradients of the normal structure, endorsing the use of skip connections to improve training. \bb{In particular, the improved PINC increased the magnitude of the gradient in the final layer by four orders of magnitude}.

\begin{figure}[tb!]
    \centering
    \includesvg[width=0.72\textwidth]{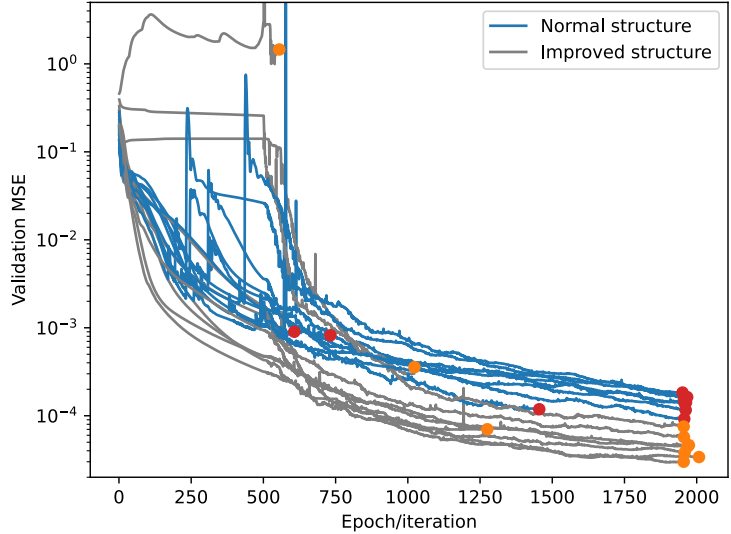}
    \caption{Results for the ablation of skip connections. Validation MSE through training of 20 neural networks, 10 with the normal dense structure and 10 with the improved structure (with skip connections). All networks were trained for 500 epochs of Adam, followed by 500 iterations of L-BFGS. 
    The larger dots indicate the endpoints of the training, either by early convergence or by completing the 500 iterations of L-BFGS. 
    Three networks of each type terminate training early. For the remainder, the improved neural network structure outperforms the standard/normal structure, 
    \bb{reducing the validation error by 67\% on average}.
    }   \label{fig:implementation_training_compare_normal_and_improved_Structure}
\end{figure}

Additionally, Figure~\ref{fig:implementation_training_compare_normal_and_improved_Structure} shows the validation error during training of 10 traditional networks in blue line and 10 networks with skip connections in gray, where the dots indicate the endpoint of the training.
All networks were trained first with 500 epochs of Adam, followed by 500 iterations of L-BFGS. 
Six neural networks, three of each type, converged before completing the 500 iterations of L-BFGS. 
  As the L-BFGS implementation evaluates the loss on average around three times per iteration, 500 iterations of L-BFGS translates to almost 2000 in the plot.
The interesting part of Figure~\ref{fig:implementation_training_compare_normal_and_improved_Structure} is that for all seven networks where the training does not stop early, the new structure with skip connections outperformed the normal structure, \bb{reducing the validation error 67\% on average}. 
In our particular and challenging application of oil well modeling, the skip connections clearly improved training.
 Further, as the average success rate for training PINC without skip connections for the oil well is quite low, we have not shown the control results for this normal architecture.
 






\section{Conclusion} \label{sec:conclusion}

This work proposed an improved version of \bb{Physics-Informed Neural Networks for Control (PINC)} of dynamic systems represented by ODEs.
  The improvements addressed a class of more realistic and complex dynamic systems, such as gas-lifted oil wells, characterized by highly nonlinear terms and functions not defined for negative numbers.
These terms restrain the training as the gradients of the physics-loss function, which contains these terms, become less informative. The extensions proposed for PINC consist of (i) using a particular neural network architecture based on skip connections, which improves the magnitude of the gradients during the training of PINCs, avoiding known gradient pathologies in PINNs \citep{improved_NN_struct}; (ii) modifying the ODE equations to avoid pitfalls for use in the loss function; and (iii) integrating a supporting feedforward neural network for learning the static mappings between system's states and algebraic variables.

\bb{Our proposed enhanced PINC achieved superior performance, averaging a 67\% reduction in the validation prediction error for the oil well application compared to the original PINC. 
Additionally, it substantially improved the gradient flow through the network layers, increasing its magnitude by four orders of magnitude.}

The results showed that the PINC framework can be successful in modeling and controlling complex systems with MPC, such as gas-lifted oil wells. PINC allowed for an open-ended long-range prediction of the system without access to the ODE (only used during training) and rendered the PINN amenable to a control scenario where the bottom-hole reference was followed. 

Future work will investigate using adaptive scaling factors $\lambda_y$ and $\lambda_F$ in the loss function, which is known to improve the training of PINNs \citep{Xiang2022}. Further, PINC could be applied to systems with other characteristics, for instance, where the reference changes continuously. The application of PINC to systems described by PDEs is also a research topic for future investigation.


\section*{Acknowledgments}
The authors acknowledge the support from UTFORSK/Norway, CAPES/Brazil (project \#88881.153841/2017-01), and CNPq/Brazil (grant \#308624/2021-1).

\input{7-Appendix.tex}

\bibliographystyle{elsarticle-harv} 
\bibliography{biblib}

\end{document}

%% file: 7-Appendix.tex
\appendix

\section{Well Model's Remaining Equations}
\label{app:well}
\input{7-Appendix_well.tex}


\section{Approximation of ODE of oil well model for PINC}
\label{app:odechange}

\begin{figure}[!ht]
    \centering
    \includesvg[width=0.7\textwidth]{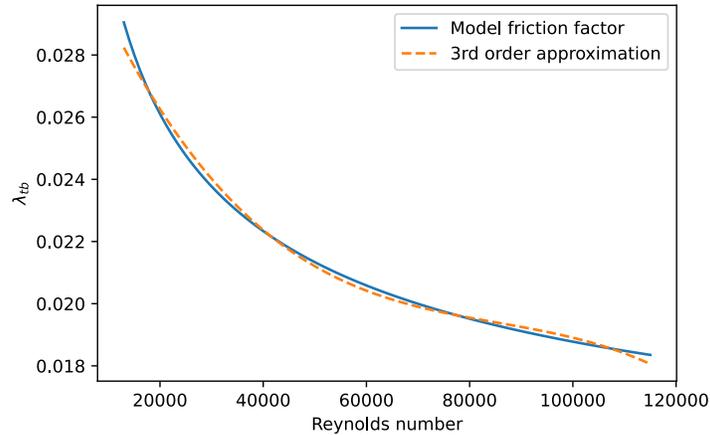}
    \caption{The third-order polynomial approximation of the tubing friction factor in Equation \eqref{eq:well_F_tb} in orange compared with the real friction factor in blue. This plot shows the approximation for the first well, which was fitted on the interval $\left [ 13000, 115000 \right ]$ using the \texttt{curve\_fit} function of SciPy.}
    \label{fig:well_friction_approx}
\end{figure}

Figure \ref{fig:well_friction_approx} shows an example of the $3^{rd}$-order polynomial approximation of the friction factor $\lambda_{\tt tb}$ as given by Equations \eqref{eq:model_approx_Re_tb} and \eqref{eq:model_approx_Re_tb_third_order_function}, in comparison to the model Equation \eqref{eq:well_alpha_L_tb_b}. 
   The range of Reynolds numbers for which the approximation is fit is: $\left [ 13000, 115000 \right ]$. 
This third-order approximation was found to resemble the original friction factor satisfactorily. It was verified by comparing Runge-Kutta simulations with the original equations and this approximation, in which there was no noticeable difference between the two.

The numerical values for the coefficients in Equation \eqref{eq:model_approx_Re_tb_third_order_function} and the ranges of Reynolds numbers in Equation \eqref{eq:model_approx_Re_tb} used to generate these approximations are shown in Table \ref{tab:implementation_3rd_order_approx_coeficients}.

\begin{table}[htb!]
\centering
\caption{Table presenting the numerical values for the third-order polynomial approximations for the three oil wells, along with the interval of Reynolds numbers for which the estimation was fitted. The bottom row shows the factor for which each column element should be multiplied, for example, for the first well: $a=-1.78\cdot 10^{-17}$.}
\label{tab:implementation_3rd_order_approx_coeficients}
\begin{tabular}{l|cccccc}
       & $a$ & $b$ & $c$ & $d$ & $Re_{{\tt tb},\min}$ & $Re_{{\tt tb},\max}$ \\ \hline
Well 1 &  $-1.78$ & $4.56$ & $-4.18$ & $3.29$  & $13000$ & $115000$ \\
Well 2 & $-1.78$ & $4.55$ & $-4.17$ & $3.29$ & $13000$ & $115000$ \\
Well 3 &  $-0.203$ &  $0.887$ & $-1.48$ &  $2.67$  & $50000$ & $160000$ \\ \hline
Factor & $10^{-17}$ & $10^{-12}$ & $10^{-7}$ & $10^{-2}$ & &  
\end{tabular}
\end{table}


\section{MPC implementation in CasADi}
\label{app:mpc}
The MPC will be formulated as a general Non-Linear Programming (NLP) problem using the \texttt{casadi.Opti\(\)} class. \bb{CasADi solves the resulting NLP problem with IPOPT \citep{IPOPT2006}. IPOPT is an interior-point algorithm developed for problems with smooth and twice differentiable objective functions and constraints, which is widely applied in NMPC and optimal control.}

In the following code listing, line 1 creates an instance of this class, then variables and parameters are added to the \texttt{Opti} object:
\begin{lstlisting}[
    caption={Instantiating variables and parameters for MPC of the bottom-hole pressure.},
    label=lst:MPC_creation,
    language=Python,
    numbers=left]
opti = casadi.Opti()
N_state=3
N_input=2
x = opti.variable(N_state, N+1)
u = opti.variable(N_input, N)
du = opti.variable(N_input, N_u)   #change in u for each step
P_bh = opti.variable(1, N)

x0 = opti.parameter(N_state,1)
u_last = opti.parameter(N_input, 1)
P_bh_ref = opti.parameter(1, N)
Q = opti.parameter((N+1), (N+1))
R = opti.parameter(N_u*N_input, N_u*N_input)
\end{lstlisting}

Here, \texttt{N\_state} is the number of states and \texttt{N\_input} is the number of control inputs (production choke and gas-lift valve). This setup is for controlling the bottom-hole pressure. The \texttt{opti.variable()} type refers to variables of the optimization problem, while \texttt{opti.parameter()} regards the parameters given to the MPC at every iteration, such as reference and weights. \texttt{x0} is the current measured (or estimated) state value, and \texttt{u\_last} is the last applied control input, which is needed as we penalize the change in control input.

An important implementation detail regards ensuring that the variables \texttt{x} and \texttt{P\_bh} satisfy the system dynamics. Code listing \ref{lst:MPC_F_mapping_x} shows that the function \texttt{opti.subject\_to()} is used only on the state variable, 
which constrains the solution to satisfy the system dynamics. 
Here, \texttt{F()} represents a trained PINC network, a function that maps a state and control input to the state 60 seconds ahead of time.
For the bottom-hole pressure, there is no need for a constraint; an assignment operator is enough. 
\texttt{F\_x\_to\_P\_bh()} represents the output of the second network trained to predict the algebraic variable bottom-hole pressure based on the values of the states and control input.
By  the \texttt{opti.subject\_to()} here as well, we add unduly complexity to the optimization problem, which results in slower computation. 
\begin{lstlisting}[
    caption={Ensuring that the MPC solution satisfy the system dynamics.},
    label=lst:MPC_F_mapping_x,
    language=Python,
    numbers=left]
opti.subject_to(x[:,0] == x0)
for i in range(0,N):            # ODE constraint
    opti.subject_to(x[:,i+1] == F(x[:,i],u[:,i])) # PINC
    
for i in range(0, N):
    P_bh[:,i] = F_x_to_P_bh(x[:,i+1], u[:,i]) # NN algebraic
\end{lstlisting}

Constraints are also needed to enforce the relationship between the control inputs $u[k]$ and the change in control inputs $\Delta u[k]$, which is implemented by the following code: 
\begin{lstlisting}[
    caption={Implementing control input change constraints.},
    label=lst:MPC_control_input,
    language=Python,
    numbers=left]
opti.subject_to(u[:,0] == u_last + du[:,0])
for i in range(0, N_u-1):
    opti.subject_to(u[:,i+1] == u[:,i] + du[:,i+1])
for i in range(N_u, N): 
    opti.subject_to(u[:,i] == u[:,N_u-1])
\end{lstlisting}

The first line ensures that the last applied control input is considered, while lines 2-3 enforce this relationship for the control input horizon. Finally, lines 4-5 fix the control input for the final steps if $N_u<N$. \texttt{du} is then used in the cost function to penalize the control input change. 

For the first simulation iteration, we need to know the current control input or set the first elements of the $\mathbf{R}$ matrix to zero. This second option allows the first control input of the MPC prediction to take any value without being penalized. This flexibility can be useful when starting a simulation from an initial condition and when a steady-state control input is unknown.


\section{Successive Linearization based MPC (SLMPC)}
\label{app:SLMPC}

Successive Linearization based MPC (SLMPC) is a controller that, at every time step, linearizes the model around the operating point and solves a linear MPC problem. This kind of controller was successfully applied to chemical reactors \citep{SLMPC_reactor} and variable stiffness actuated robots \citep{SLMPC_robots}. 

At every iteration of the MPC, the system equations are linearized and discretized around the current operating point $\mathbf{x}_k$, current control input $\mathbf{u}_k$, and current control variable $\mathbf{y}_k$. The linearized model can then be written as:
\begin{subequations}
\begin{align}
    \Delta\mathbf{x}_{k+j+1} &= \Delta\mathbf{x}_{k+j} + \mathbf{A}_k\Delta\mathbf{x}_{k+j} + \mathbf{B}_k\Delta\mathbf{u}_{k+j} + \mathbf{\delta}_k \label{eq:implementation_SLMPC_F_mapping}\\
    \mathbf{y}_{k+j} &= \mathbf{C}_k\Delta\mathbf{x}_{k+j} + \mathbf{D}_k\Delta\mathbf{u}_{k+j} + \mathbf{y}_k  \label{eq:implementation_SLMPC_Fy_mapping} \\
    \Delta\mathbf{x}_{k+j} &= \mathbf{x}_{k+j} - \mathbf{x}_k, \quad \Delta\mathbf{u}_{k+j} = \mathbf{u}_{k+j} - \mathbf{u}_k,
\end{align}
\label{eq:implementation_SLMPC}
\end{subequations}
where:
\begin{subequations}
\begin{align}
    &\mathbf{A}_k = T\frac{\partial \mathbf{f}}{\partial \mathbf{x}_k}(\mathbf{x}_k, \mathbf{u}_k), \,
    \mathbf{B}_k = T\frac{\partial \mathbf{f}}{\partial \mathbf{u}_k}(\mathbf{x}_k, \mathbf{u}_k), \\
    &    \mathbf{\delta}_k = T\mathbf{f}(\mathbf{x}_k, \mathbf{u}_k),\\
    &\mathbf{C}_k = \frac{\partial \mathbf{h}}{\partial \mathbf{x}_k}(\mathbf{x}_k, \mathbf{u}_k), \,
    \mathbf{D}_k = \frac{\partial \mathbf{h}}{\partial \mathbf{u}_k}(\mathbf{x}_k, \mathbf{u}_k), \,
    \mathbf{y}_k  =\mathbf{h}(\mathbf{x}_k, \mathbf{u}_k),
\end{align}
\label{eq:implementation_SLMPC_Matrices}
\end{subequations}
\noindent where $T$ is the discretization time step length, $\mathbf{f}(\mathbf{x}, \mathbf{u})$ is the ODE system, and $\mathbf{h}$ is a function that computes the output from the states and control input: $\mathbf{y}=\mathbf{h}(\mathbf{x}, \mathbf{u})$. Then, the linear model in Equation \eqref{eq:implementation_SLMPC_F_mapping} can replace the PINC in the constraint of Equation \eqref{eq:backgorund_NMPC_y_F}, and the linearized output Equation \eqref{eq:implementation_SLMPC_Fy_mapping} can be used to implement the constraint in Equation \eqref{eq:backgorund_NMPC_y_Fy}.


%% file: 7-Appendix_well.tex
\subsection{Parameters of oil well models}

Table \ref{tab:model_constants} shows the model parameters used for the primary oil well, which were extracted from \cite{well_model_paper:2012}, except for $\epsilon$ which could not be found in the article. Instead, it was collected from \cite{dissertation_jean}. Several oil wells were constructed by altering the parameters of the model, whose values are shown in Table \ref{tab:model_constants_all_wells}.

\begin{table*}[tb!]
\caption{Parameters for the oil well model from \cite{well_model_paper:2012}. }
    \label{tab:model_constants}
$$
\begin{array}{cccc}
\hline 
  \text { Symb. } & \text { Description } & \text { Values } & \text { Units } \\
\hline 
R & \text { universal gas constant } & 8.314 & \mathrm{~J} /(\mathrm{mol} \cdot \mathrm{K}) \\
g & \text { gravity } & 9.81 & \mathrm{~m} / \mathrm{s}^{2} \\
\mu & \text { viscosity } & 3.64 \times 10^{-3} & \mathrm{~Pa} \cdot \mathrm{s} \\
\rho_{\tt L} & \text { liquid density } & 760 & \mathrm{~kg} / \mathrm{m}^{3} \\
M_{\tt G} & \text { gas molecular weight } & 0.0167 & \mathrm{kg} / \mathrm{mol} \\
T_{\tt an} & \text { annulus temperature } & 348 & \mathrm{~K} \\
V_{\tt an} & \text { annulus volume } & 64.34 & \mathrm{~m}^{3} \\
L_{\tt an} & \text { annulus length } & 2048 & \mathrm{~m}^{3} \\
P_{\tt g s} & \text { gas source pressure } & 140 & \mathrm{bar} \\
S_{\tt bh} & \text { cross-section below injection point } & 0.0314 & \mathrm{~m}^{2} \\
L_{\tt bh} & \text { length below injection point } & 75 & \mathrm{~m} \\
T_{\tt tb} & \text { injection point tubing temperature } & 369.4 & \mathrm{~K} \\
G O R & \text { mass gas oil ratio } & 0 & - \\
P_{\tt r e s} & \text { reservoir pressure } & 160 & \mathrm{bar} \\
\overline{w}_{\tt r e s} & \text { average mass flow from reservoir } & 18 & \mathrm{~kg} / \mathrm{s} \\
D_{\tt tb} & \text { tubing diameter } & 0.134 & \mathrm{~m} \\
L_{\tt tb} & \text {  tubing length }  & 2048 & \mathrm{~m} \\
V_{\tt tb} & \text { tubing volume } & 25.03 & \mathrm{~m}^{3} \\
\epsilon & \text{ piping superficial roughness } & 2.80 \times 10^{-5} & m \\ 
P I & \text {  productivity index } & 2.47 \times 10^{-6} & \mathrm{~kg} /(\mathrm{s \cdot Pa}) \\
K_{\tt gs} & \text { gas-lift choke cons. } & 9.98 \times 10^{-5} & - \\
K_{\tt inj} & \text { injection valve cons. } & 1.40 \times 10^{-4} & - \\
K_{\tt pr} & \text {  production choke cons. } & 2.90 \times 10^{-3} & - \\
\hline
\end{array}
$$

\end{table*}


\begin{table}[tb!]
$$
\begin{array}{cccc}
\hline \text { Symb. } & \text { Well 1 } & \text { Well 2 } & \text { Well 3 } \\
\hline R &8.314 &8.314 &8.314 \\
g &9.81 &9.81 &9.81 \\
\mu &3.64 \times 10^{-3} &3.64 \times 10^{-3} &3.64 \times 10^{-3} \\
\rho_{L} &760 &760 &730 \\
M_{G} &0.0167 &0.0167 &0.0167 \\
T_{an} &348 &335 &360 \\
V_{an} &64.34 &84.82 &56.55 \\
L_{an} &2048 &2700 &1800 \\
P_{g s} &140 &140 &140 \\
S_{bh} &0.0314 &0.0314 &0.0314 \\
L_{bh} &75 &75 &40 \\
T_{tb} &369.4 &355.6 &381.2 \\
G O R &0 &0 &0.2 \\
P_{r e s} &160 &165 &157 \\
\overline{w}_{r e s} &18 &11 & 30 \\
D_{tb} &0.134 &0.130 &0.134 \\
L_{tb} &2048 &2700 &1800 \\
V_{tb} &25.03 &31.00 &22.08 \\
\epsilon &2.80 \times 10^{-5} &2.80 \times 10^{-5} &2.80 \times 10^{-5} \\
P I &2.47 \times 10^{-6} &2.12 \times 10^{-6} &3.89 \times 10^{-6} \\
K_{g s} &9.98 \times 10^{-5} &10.43 \times 10^{-5} &3.89 \times 10^{-5} \\
K_{i n j} &1.40 \times 10^{-4} &1.20 \times 10^{-4} &1.78 \times 10^{-4} \\
K_{p r} &2.90 \times 10^{-3} &2.43 \times 10^{-3} &3.22 \times 10^{-3} \\
\hline
\end{array}
$$
\caption{Parameters for all the wells. First column is equivalent to the model from Table \ref{tab:model_constants} }
    \label{tab:model_constants_all_wells}
\end{table}


\subsection{Mass flows}

\begin{subequations}
\begin{align}
    w_{{\tt G},{\tt in}} &= K_{{\tt gs}} u_2 \sqrt{\rho_{{\tt G},{\tt in}} \max(P_{\tt gs} - P_{\tt at}, 0)} \label{eq:well_w_G_in} \\
    w_{{\tt G},{\tt inj}} &= K_{\tt inj} \sqrt{\rho_{{\tt G},{\tt an},{\tt b}} \max(P_{{\tt an},{\tt b}} - P_{{\tt tb},{\tt b}}, 0)} \label{eq:well_w_G_inj} \\
    w_{\tt res} &= PI \max(P_{\tt res} - P_{\tt bh}, 0) \label{eq:well_w_res} \\
    w_{{\tt L},{\tt res}} &= (1- \alpha^{\tt m}_{{\tt G},{\tt tb},{\tt b}}) w_{\tt res} \label{eq:well_w_L_res} \\
    w_{{\tt G},{\tt res}} &= \alpha^{\tt m}_{{\tt G},{\tt tb},{\tt b}} w_{\tt res} \label{eq:well_w_G_res} \\
    w_{\tt out} &= K_{\tt pr} u_1 \sqrt{\rho_{{\tt mix},{\tt tb},{\tt t}} \max(P_{{\tt tb},{\tt t}}-P_0, 0)} \label{eq:well_w_out} \\
    w_{{\tt L},{\tt out}} &= (1 - \alpha^{\tt m}_{{\tt G},{\tt tb},{\tt t}}) w_{\tt out} \label{eq:well_w_L_out} \\
    w_{{\tt G},{\tt out}} &= \alpha^{\tt m}_{{\tt G},{\tt tb},{\tt t}} w_{\tt out} \label{eq:well_w_G_out}
\end{align}
\label{eq:well_mass_flows}
\end{subequations}

The gas flow entering the annulus is $w_{{\tt G},{\tt in}}$ at the top, while $w_{{\tt G},{\tt inj}}$ is the flow of gas leaving the annulus and  entering the production tubing. 
   The total mass flow, liquid flow, and gas flow entering the production tubing from the reservoir are given by $w_{\tt res}$, $w_{{\tt L},{\tt res}}$, and $w_{{\tt G},{\tt res}}$ respectively.
The total mass flow, liquid flow, and gas flow leaving the production tubing at the top-side are $w_{\tt out}$, $w_{{\tt L},{\tt out}}$, and $w_{{\tt G},{\tt out}}$ respectively.
   
\subsection{Pressures}

\begin{subequations}
\begin{align}
    P_{{\tt an},{\tt t}} &= \frac{R T_{\tt a} m_{{\tt G},{\tt an}}}{M_{\tt G} V_{\tt a}} \label{eq:well_P_an_t} \\
    P_{{\tt an},{\tt b}} &= P_{{\tt an},{\tt t}} + \frac{m_{{\tt G},{\tt an}} g L_{\tt an}}{V_{\tt an}} \label{eq:well_P_an_b} \\
    P_{{\tt tb},{\tt t}} &= \frac{\rho_{{\tt G},{\tt tb},{\tt t}} R T_{\tt tb}}{M_{\tt G}} \label{eq:well_P_tb_t} \\
    P_{{\tt tb},{\tt b}} &= P_{{\tt tb},{\tt t}} + \overline{\rho}_{\tt mix} g L_{\tt tb} + F_{\tt tb} \label{eq:well_P_tb_b} \\
    P_{\tt bh} &= P_{{\tt tb},{\tt b}} + F_{\tt bh} + \rho_{\tt L} g L_{\tt bh} \label{eq:well_P_bh}
\end{align}
\label{eq:well_pressures}
\end{subequations}

The equations above define the pressure $P_{{\tt an},{\tt t}}$ at the top  and the pressure $P_{{\tt an},{\tt b}}$ at the bottom of the annulus. Likewise, $P_{{\tt tb},{\tt t}}$ and $P_{{\tt tb},{\tt b}}$ correspond to the pressure at the top and bottom of the tubing, respectively.
  Finally, $P_{\tt bh}$  is the bottom-hole pressure.

\subsection{Densities}

The densities of gas and liquid in the streams in the annulus and tubing, when applicable, are given by the equations below.
\begin{subequations}
\begin{align}
    \rho_{{\tt G},{\tt an},{\tt b}} &= \frac{P_{{\tt an},{\tt b}} M_{\tt G}}{R T_{\tt an}} \label{eq:well_rho_G_an_b} \\
    \rho_{{\tt G},{\tt in}} &= \frac{P_{\tt gs} M_{\tt G}}{R T_{\tt an}} \label{eq:well_rho_G_in} \\
    \rho_{{\tt G},{\tt tb},{\tt t}} &= \frac{m_{{\tt G},{\tt tb}}}{V_{\tt tb} + S_{\tt bh} L_{\tt bh} - m_{{\tt L},{\tt tb}}/\rho_{\tt L}} \label{eq:well_rho_G_tb_t} \\
    \overline{\rho}_{\tt mix} &= \frac{m_{{\tt G},{\tt tb}} + m_{{\tt L},{\tt tb}} - \rho_{\tt L} S_{\tt bh} L_{\tt bh}}{V_{\tt tb}} \label{eq:well_rho_mix} \\
    \rho_{{\tt G},{\tt tb},{\tt b}} &= \frac{P_{{\tt tb},{\tt b}} M_G}{R T_{\tt tb}} \label{eq:well_rho_G_tb_b} \\
    \rho_{{\tt mix},{\tt tb},{\tt t}} &= \alpha_{{\tt L},{\tt tb},{\tt t}} \rho_{\tt L} + (1-\alpha_{{\tt L},{\tt tb},{\tt t}}) \rho_{{\tt G},{\tt tb},{\tt t}} \label{eq:well_rho_mix_tb_t}
\end{align}
\label{eq:well_densities}
\end{subequations}

\subsection{Mass/liquid fractions}
\begin{subequations}
\begin{align}
    \overline{\alpha}_{{\tt L},{\tt tb}} &= \frac{m_{{\tt L},{\tt tb}} - \rho_{\tt L} S_{\tt bh} L_{\tt bh}}{V_{\tt tb} \rho_{\tt L}} \label{eq:well_alpha_L_tb} \\
    \alpha^{\tt m}_{{\tt G},{\tt bh}} &= GOR/(GOR+1) \label{eq:well_alpha_G_tb_b} \\
    \alpha_{{\tt L},{\tt tb},{\tt b}} &= \frac{w_{{\tt L},{\tt res}} \rho_{{\tt G},{\tt tb},{\tt b}}}{w_{{\tt L},{\tt res}} \rho_{{\tt G},{\tt tb},{\tt b}} + (w_{{\tt G},{\tt inj}} + w_{{\tt G},{\tt res}})\rho_{\tt L}} \label{eq:well_alpha_L_tb_b} \\
    \alpha_{{\tt L},{\tt tb},{\tt t}} &= 2 \overline{\alpha}_{{\tt L},{\tt tb}} - \alpha_{{\tt L},{\tt tb},{\tt b}} \label{eq:well_alpha_L_tb_t} \\
    \alpha^{\tt m}_{{\tt G},{\tt tb},{\tt t}} &= \frac{(1 - \alpha_{{\tt L},{\tt tb},{\tt t}}) \rho_{{\tt G},{\tt tb},{\tt t}}}{\alpha_{{\tt L},{\tt tb},{\tt t}} \rho_{\tt L} + (1 - \alpha_{{\tt L},{\tt tb},{\tt t}}) \rho_{{\tt G},{\tt tb},{\tt t}}} \label{eq:well_alpha_G_tb_t}
\end{align}
\label{eq:well_mass_liquid_fractions}
\end{subequations}

\subsection{Velocities}\label{sec:sub_well_veloctiries}

\begin{subequations}
\begin{align}
    \overline{U}_{{\tt L},{\tt tb}} &= \frac{4(1 - \alpha^m_{{\tt G},{\tt bh}}) \overline{w}_{\tt res}}{\rho_{\tt L} \pi D^2_{\tt tb}} \label{eq:well_U_L_tb} \\
    \overline{U}_{{\tt G},{\tt tb}} &= \frac{4(w_{{\tt G},{\tt in}} + \alpha^{\tt m}_{{\tt G},{\tt bh}} \overline{w}_{\tt res})}{\rho_{{\tt G},{\tt tb},{\tt t}} \pi D^2_{\tt tb}} \label{eq:well_U_G_tb} \\
    \overline{U}_{{\tt mix},{\tt tb}} &= \overline{U}_{{\tt L},{\tt tb}} + \overline{U}_{{\tt G},{\tt tb}} \label{eq:well_U_mix_tb} \\
    \overline{U}_{{\tt L},{\tt bh}} &= \frac{\overline{w}_{\tt res}}{\rho_L S_{\tt bh}} \label{eq:well_U_L_bh} 
\end{align}
\label{eq:well_velocities}
\end{subequations}

These equations express the liquid, gas and mix velocities in the tubing, and the liquid velocity in the bottom-hole. 

\subsection{Friction terms}

\begin{subequations}
\begin{align}
    Re_{\tt tb} &= \frac{\overline{\rho}_{{\tt mix},{\tt tb}} \overline{U}_{{\tt mix},{\tt tb}} D_{\tt tb}}{\mu} \label{eq:well_Re_tb} \\
    \frac{1}{\sqrt{\lambda_{\tt tb}}} &= -1.8\log_{10} \left[ \left(\frac{\epsilon / D_{\tt tb}}{3.7}\right)^{1.11} + \frac{6.9}{Re_{\tt tb}} \right] \label{eq:well_lambda_tb} \\
    F_{\tt tb} &= \frac{\overline{\alpha}_{{\tt L},{\tt tb}} \lambda_{\tt tb} \overline{\rho}_{{\tt mix},{\tt tb}} \overline{U}^2_{{\tt mix},{\tt tb}} L_{{\tt tb}}}{2 D_{\tt tb}} \label{eq:well_F_tb} \\
     Re_{\tt bh} &= \frac{\rho_{\tt L} \overline{U}_{{\tt L},{\tt bh}} D_{\tt bh}}{\mu} \label{eq:well_Re_h} \\
     \frac{1}{\sqrt{\lambda_{\tt bh}}} &= -1.8\log_{10} \left[ \left(\frac{\epsilon / D_{bh}}{3.7}\right)^{1.11} + \frac{6.9}{Re_{\tt bh}} \right] \label{eq:well_lambda_bh} \\
     F_{\tt bh} &= \frac{\lambda_{\tt bh} \rho_{\tt L} \overline{U}^2_{{\tt L},{\tt bh}} L_{\tt bh}}{2 D_{\tt bh}} \label{eq:well_F_bh}
\end{align}
\label{eq:well_friction_terms}
\end{subequations}

These equations calculate the friction pressure-loss terms for the tubing \eqref{eq:well_F_tb} and the bottom-hole \eqref{eq:well_F_bh}. For the bottom-hole friction calculations, the average mass flow from the reservoir, $\overline{w}_{res}$, is used, as can be seen in Equation \eqref{eq:well_U_L_bh}, resulting in Equations \eqref{eq:well_Re_h}, \eqref{eq:well_lambda_bh} and \eqref{eq:well_F_bh} being constant, which is another of the simplifications made in the derivation of the model to make it explicit.

\subsection{Simplifications and Discussion}

Simplifications were implemented in the algebraic equations in order to avoid implicit terms, resulting in an explicit model in the form of a system of ODE equations that can be simulated with methods such as Runge-Kutta.

Equations \eqref{eq:well_U_L_tb}, \eqref{eq:well_U_G_tb} and \eqref{eq:well_U_L_bh} use the predefined constant $\overline{w}_{\tt res}$ as the flow from the reservoir, which is one of the simplifications in the model derived in \cite{well_model_paper:2012} to avoid an implicit set of equations. 

For the bottom-hole friction calculations, the average mass flow from the reservoir, $\overline{w}_{res}$, is used, as can be seen in Equation \eqref{eq:well_U_L_bh}, resulting in Equations \eqref{eq:well_Re_h}, \eqref{eq:well_lambda_bh} and \eqref{eq:well_F_bh} being constant, which is another simplification in the derivation of the model to render it explicit. 

Equation \eqref{eq:well_U_G_tb} relates the velocity of gas to the mass flow of gas injected into the tubing ($w_{{\tt G},{\tt inj}}$), but the latter is replaced with the mass flow of gas injected into the annulus through the gas-lift choke ($w_{{\tt G},{\tt in}}$). This allows changes in the gas-lift choke opening to directly affect the velocities in the tubing, the tubing friction loss, and the pressure at the bottom of the tubing, which happens more quickly than the change in pressure at the bottom of the annulus. The mass flow of gas injected into the tubing $w_{{\tt G},{\tt inj}}$ is dependent on the pressure difference between the bottom of the annulus and the bottom of the tubing. A change in the gas-lift choke opening has an immediate effect on the gas injected into the tubing.

Table \ref{tab:model_constants} shows the model parameters of the primary oil well, which were
extracted from \cite{well_model_paper:2012}, except for $\epsilon$ which could not be found in the article and instead it was collected from \cite{dissertation_jean}.
